\pdfoutput=1

\documentclass[11pt]{article}

\usepackage[final]{acl}

\usepackage{tikz,tikz-dependency}
\usepackage{times}
\usepackage{latexsym}
\usepackage{amssymb}
\usepackage{mathtools}
\usepackage{multirow}
\usepackage{subcaption}

\usepackage[T1]{fontenc}

\usepackage[utf8]{inputenc}

\usepackage{microtype}

\usepackage{inconsolata}

\usepackage{graphicx}
\usepackage{booktabs}
\usepackage{amsmath}
\usepackage{bbold}
\usepackage[inkscapelatex=false]{svg}
\usepackage{placeins}
\usepackage{float}
\usepackage{listings}
\usepackage{xcolor}
\usepackage{courier}
\usepackage{cancel}

\newcommand{\glove}{GloVe}

\newcommand{\model}{\ensuremath{f_\theta}}
\newcommand{\probe}{\ensuremath{g_\phi}}
\newcommand{\struct}{\ensuremath{g^\text{struct}_{\phi}}}
\newcommand{\ortho}{\ensuremath{g^\text{ortho}_{\phi}}}
\newcommand{\head}{\ensuremath{g^\text{head}_{\phi}}}
\newcommand{\control}{\ensuremath{g^\text{ctrl}_{\phi}}}

\newcommand{\dprobe}{\ensuremath{k}}
\newcommand{\dmodel}{\ensuremath{n}}

\definecolor{mygreen}{RGB}{40,200,40}

\title{Mechanisms vs.\ Outcomes: Probing for Syntax Fails to Explain Performance on Targeted Syntactic Evaluations}

\author{Ananth Agarwal, Jasper Jian, Christopher D. Manning, Shikhar Murty \\
        Stanford University \\ \texttt{\{ananthag, manning, smurty\}@cs.stanford.edu, jjian@stanford.edu}}

\begin{document}
\maketitle
\begin{abstract}
Large Language Models (LLMs) exhibit a robust mastery of syntax when processing and generating text. While this suggests internalized understanding of hierarchical syntax and dependency relations, the precise mechanism by which they represent syntactic structure is an open area within interpretability research. Probing provides one way to identify syntactic mechanisms linearly encoded in activations; however, no comprehensive study has yet established whether a model's probing accuracy reliably predicts its downstream syntactic performance. Adopting a ``mechanisms vs.\ outcomes'' framework, we evaluate 32 open-weight transformer models and find that syntactic features extracted via probing fail to predict outcomes of targeted syntax evaluations across English linguistic phenomena. Our results highlight a substantial disconnect between latent syntactic representations found via probing and observable syntactic behaviors in downstream tasks.
\end{abstract}

\section{Introduction}
The remarkable ability of LLMs to process and generate text that respects a rich diversity of syntactic constraints strongly suggests that models have sophisticated syntactic knowledge. Yet our understanding of internal representations of these syntactic facts is lacking, which has motivated research into interpretability paradigms searching for syntactic \textit{mechanisms}, such as mechanistic interpretability and causal abstractions \cite{geiger2022, murty2023intrinsic}. We study probing, a prominent approach in this domain which assumes that model \textit{activations} are the mechanism through which syntactic knowledge is encoded and that this knowledge is linearly recoverable using small supervised models called \textit{probes} \citep[among others]{conneau-etal-2018-cram, hewitt-manning-2019-structural, tenney-etal-2019-bert}. However, no comprehensive study has evaluated whether probing accuracy is indicative of targeted syntactic \textit{outcomes}.

\begin{figure}
\centering
\includegraphics[width=0.9\columnwidth]{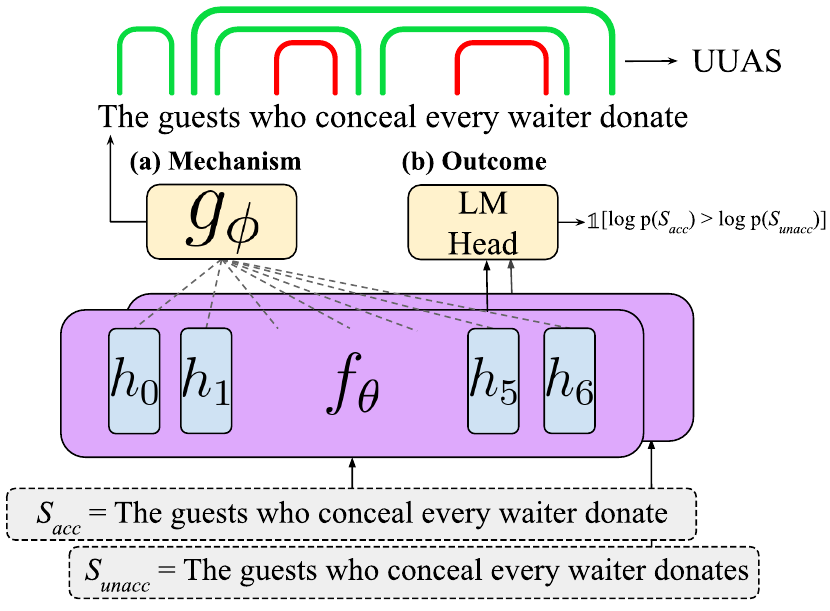}
    \caption{\textbf{Mechanisms vs. outcomes setup}. We find no convincing predictive power of syntax probing accuracy on downstream syntactic evaluation accuracy. (a) Mechanism: probe \probe{} extracts a dependency parse tree from $S_{acc}$ word-level hidden states $h_i$; UUAS = 4/6 edges in this toy example. (b) Outcome: evaluate minimal pair accuracy.} 
  \label{fig:fig_1_design}
\end{figure}

We deepen understanding of the relationship between syntactic mechanisms and outcomes by asking: \textit{Do probing accuracies, measured by dependency tree attachment scores of syntactic probes, effectively predict a model's downstream performance on targeted syntactic evaluations?} We evaluate syntactic outcomes using BLiMP \citep{warstadt-etal-2020-blimp-benchmark}, a benchmark of minimal pairs spanning well-established English grammatical contrasts. Assigning higher probability to the acceptable sentence in an acceptable-unacceptable pair has been established as a desired linguistic outcome \cite{marvin-linzen-2018-targeted, warstadt-etal-2020-blimp-benchmark}.

We train three syntax probes on 32 open-weight transformer models. We then fit Ordinary Least Squares (OLS) regressions modeling BLiMP minimal pairs accuracy as a function of probe attachment scores on the grammatically acceptable sentences (Fig~\ref{fig:fig_1_design}). We demonstrate that across all levels of analysis---amalgamation of all BLiMP linguistic phenomena, individual phenomena, and targeted suites of difficult minimal pairs---syntax probing accuracy shows no clear predictive power for downstream outcomes.\footnote{Code is at \url{https://github.com/agananth/SyntaxMechanismsOutcomes}} 

Concretely, we find that probing performance—measured via directed and undirected unlabeled attachment scores (UAS/UUAS)—fails to predict model accuracy on BLiMP minimal pairs. Across 32 models, none of the three probes yields a statistically significant regression fit on the full overall dataset. At the per-phenomenon level, only anaphor agreement shows a significant correlation, but the control experiment reveals that even a non-syntactic probe can produce a similar fit, suggesting the result is likely spurious. In finer-grained evaluations targeting difficult syntactic paradigms, we test whether probes can recover critical tree edges that are plausibly relevant for resolving minimal pairs, such as the subject-verb edge in subject--verb agreement pairs. However, probe recovery of these edges only weakly aligns with minimal pair outcome (40–60\% match rate across models). Finally, within each BLiMP paradigm we test whether UUAS score distributions differ between correct and incorrect minimal pair outcomes. In over 85\% of paradigms, the distributions overlap substantially for most models.

Taken together, our results suggest that while syntactic probing is a dominant interpretability paradigm, it fails to robustly reveal a model’s latent syntactic knowledge. We establish a disconnect between the syntactic information revealed through conventional probing for dependency trees and the latent syntactic knowledge that far more often than not assigns grammatical text a higher likelihood than ungrammatical text. Our results argue for the use of external targeted evaluations as the gold standard for establishing model syntactic competence, and support the continuing development of BLiMP-style resources beyond English \citep[among others]{jumelet2025multiblimp10massivelymultilingual, başar2025turblimpturkishbenchmarklinguistic, taktasheva2024rublimprussianbenchmarklinguistic}. 

\section{Setup}

Given an input sentence $S \triangleq (w_1, \ldots, w_N)$, a pre-trained transformer \model{} produces contextual vectors $h_i \in \mathbb{R}^{\dmodel{}}$ for each word $w_i$ at every layer. Our probes, denoted $\probe : \mathbb{R}^{\dmodel{}} \to \mathbb{R}^{\dprobe{}}$ with parameters $\phi$, are learned linear functions that project contextual hidden states into a $\dprobe{}$-dimensional subspace, where $\dprobe{} \leq \dmodel{}$. Linear and bilinear probes have been shown to achieve high selectivity; they effectively capture linguistic properties of the probed representation \citep{hewitt2019designinginterpretingprobescontrol}.

\subsection{Mechanisms}

We train three syntax probes with different optimization objectives for comprehensive coverage. We evaluate the correctness with which each probe extracts the dependency parse of a sentence.

\paragraph{Structural Probe.} Proposed in \citet{hewitt-manning-2019-structural}, the structural distance probe \struct{} is a linear transformation $\struct(h) = B^{\text{struct}}h$ with parameters $\phi = \{B^\text{struct} \in \mathbb{R}^{\dprobe{} \times \dmodel{}} \}$ that learns to encode the dependency tree distance $d_{ij}$ between word pair $w_i$ and $w_j$ in the probe's projected space by minimizing 
\begin{gather} \label{eq:}
\min_{\phi} \sum_{S} \frac{1}{|S|^2} \sum_{i, j} \left| d_{ij} - \|\struct (h_i - h_j)\|_2^2\right|.    
\end{gather}

The evaluation metric is undirected unlabeled attachment score (UUAS): the fraction of gold tree edges $E_\text{gold} = \{\{w_i,w_j\} : (w_i,w_j) \in \mathcal{G}\}$ included in the minimum spanning tree $E_\text{pred} = \{\{w_i,w_j\} : (w_i,w_j) \in \hat{\mathcal{G}}\}$ calculated from the probe's predicted tree distances. Punctuation and the root relation are excluded.
\begin{gather*}
\text{UUAS} = (|E_\text{pred} \cap E_\text{gold}|) / |E_\text{gold}|.
\end{gather*}

\paragraph{Orthogonal Structural Probe.} \citet{Limisiewicz2021orth} replace the original structural probe linear transformation with an orthogonal transformation and scaling vector. The orthogonal probe \ortho{} is the transformation $\ortho(h) = \bar{d} \odot Vh$ with parameters $\phi = \{V \in \mathbb{R}^{\dmodel {} \times \dmodel}, \bar{d} \in \mathbb{R}^\dmodel \}$ where $V$ is orthogonal. The authors maintain orthogonality during training using Double Soft Orthogonality Regularization (DSO; \citealp{bansal2018gainorthogonalityregularizationstraining}) with regularization $\lambda_o = 0.05$. The overall probe training objective is

\begin{gather}
\min_{\phi} \sum_{S} \frac{1}{|S|^2} \sum_{i, j} \left| d_{ij} - \|\ortho (h_i - h_j)\|_2^2\right| \nonumber \\ \phantom{\min_{\phi} \sum_{S} \frac{1}{|S|^2} \sum_{i, j} d_{ij} - \ortho } + \lambda_o \text{DSO}(V).    
\end{gather}

\citet{Limisiewicz2021orth} report \ortho{} performs on par with \struct{} at predicting dependency trees while being less prone to memorizing training trees. Moreover, the scaling vector $\bar{d}$ enables interpretation of the relative importance of each dimension in $V$. The evaluation metric is UUAS, as with \struct{}.

\paragraph{Head Word Probe.} Inspired by \citet{clark2019bert}, we define the head word probe \head{} as a linear transformation $\head(h) = B^\text{head}h$ with parameters $\phi = \{B^\text{head} \in \mathbb{R}^{\dprobe{} \times \dmodel{}} \}$ trained using cross-entropy loss. For each dependent $w_i$ (including punctuation), the probe predicts its head in $\{\text{ROOT}, w_1, \ldots, w_N\} \setminus \{w_i\}$. Let $H$ be cross entropy loss and $\text{head}(i)$ be the head of $w_i$:
\begin{gather}
    \hat{d}_{ij} = \| \head(h_i - h_j) \|_2 \nonumber \\ 
    \min_{\phi} \sum_{S} \frac{1}{|S|} \sum_i H(\{\hat{d}_{ij} \mid i \neq j\}, \text{head}(i)). \label{eq:head}
\end{gather}
Since the edge prediction is directed, the evaluation metric is unlabeled attachment score (UAS). Here $E_\text{gold} = \{(i, \mathrm{head}(i)) : 1 \leq i \leq N \}$ and $E_\text{pred} = \{(i, \hat{\mathrm{head}}(i)) : 1 \leq i \leq N \}$ such that
\begin{gather*}
\text{UAS} = (|E_\text{pred} \cap E_\text{gold}|) / |E_\text{gold}|.
\end{gather*}

\paragraph{Control Probe.}
\label{sec:control}
Importantly, the performance of syntax probes and downstream syntactic performance may partly reflect non-syntactic factors such as the quality of lexical semantic representations in a model. The overall quality of model hidden states is thus a confounding variable affecting both probe and syntactic evaluation performance. To isolate this confounder, we propose a simple control probing task where performance depends only on hidden state quality and not latent syntactic structure. Concretely, for each sentence, we form all word pairs where the treebank-specific part-of-speech (XPOS) of the two words is the same and train a probe \control{} to recover the $L_2$ distance between the two words' \glove{} vectors \cite{pennington-etal-2014-glove}. \control{} is a linear transformation $\control(h) = B^\text{control}h$ with parameters $\phi = \{B^\text{control} \in \mathbb{R}^{\dprobe{} \times \dmodel{}} \}$. Let $v_i$ denote the \glove{} vector of $w_i$. The training objective for the control probe is
\begin{gather} 
    \tilde{d}_{ij} = \| \control(h_i - h_j) \|_2 \nonumber \\ 
    \min_{\phi} \sum_S \frac{1}{|S|}\sum_{\mathclap{\substack{(i, j) \\ \text{XPOS}(w_i) = \text{XPOS}(w_j)}}}\; \text{Huber}(\tilde{d}_{ij}, \|v_i - v_j\|_2). \label{eq:control}
\end{gather}
We use Huber loss for outlier robustness. The evaluation metric for the probe is the Spearman correlation $\rho_s$ between predicted and actual \glove{} distances. 

This probe is trained to recover non-syntactic information as modeled by an uncontextualized word embedding model, \glove{}, whose vector geometries have been shown to encode linguistic information like lexical semantics \citep{https://doi.org/10.1111/cogs.13291}. We further ensure that our control probe is not indirectly capturing syntactic part-of-speech information by only training the probe to recover within word-pair distances where the words share the same XPOS. Crucially, any predictive success of the control probe on syntactic evaluations warrants a more cautious interpretation of the results of syntactic probes, as it suggests that the latter's explanatory power may not be attributable solely to the recoverability of syntactic representations.

\subsection{Outcomes}
Our benchmark for linguistic knowledge outcomes is BLiMP accuracy. BLiMP has 67 template-generated sub-datasets called paradigms grouped into 13 linguistic phenomena across English morphology, syntax, and semantics \cite{warstadt-etal-2020-blimp-benchmark}.\footnote{The two semantic phenomena, quantifiers and NPI licensing, still require knowledge of hierarchical syntactic structure, e.g., to determine scope for NPI licensing \citep{Ladusaw1979}.} Each paradigm has 1000 acceptable-unacceptable minimal pairs $(S_{acc}, S_{unacc})$ that exhibit a single grammatical contrast. The example pair from Fig~\ref{fig:fig_1_design} is from the \textit{distractor agreement relative clause} paradigm within subject--verb agreement. The verb agreement contrast point in the pair between the acceptable (top) and unacceptable (bottom) sentences is shown below in bold: 
\[
\underbrace{\text{The guests }}_{\text{Subject (plural)}} \underbrace{\text{who conceal } \underbrace{\text{every waiter}}_{\text{Distractor (singular)}}}_{\text{Relative Clause}} 
\text{ \textbf{\textcolor{mygreen}{donate}}.}
\]

\[
\underbrace{\text{The guests }}_{\text{Subject (plural)}} \underbrace{\text{who conceal } \underbrace{\text{every waiter}}_{\text{Distractor (singular)}}}_{\text{Relative Clause}} \text{ \textbf{\textcolor{red}{donates}}.}
\]

Model \model{} accuracy on a set of pairs $\mathcal{D}$ is computed as
\begin{equation}
\label{eq:blimp}
    \frac{1}{|\mathcal{D}|} \sum_{\mathcal{D}} \mathbb{1} \{\log P_{\model}(S_{acc}) > \log P_{\model}(S_{unacc}) \}.
\end{equation}
For decoder models, $\log P_{\model}(S)$ is the sum of token level log-probabilities. For encoder and encoder-decoder models, we instead use pseudo-log-likelihood scores (PLL; \citealp{Salazar_2020}). 

\subsection{Models}
We test 32 open-weight pretrained decoder, encoder, and encoder-decoder models up to 8B parameters. Appendix~\ref{sec:full-model-list} has the full list.

\section{Experimental Setup}\label{sec:exp-setup}

\paragraph{Probe Training.}
We train each of our four probe ``families'' on Stanza \cite{qi-etal-2020-stanza} Universal Dependencies \cite{de-marneffe-etal-2021-universal} dependency parses of the Penn Treebank (PTB; \citealp{marcus-etal-1993-building}) using its standard train and validation splits. We report training parameters and compute in Appendix~\ref{sec:probe-training}. \struct{}, \head{}, and \control{} have fixed probe dimension $\dprobe{} = 256$, while \ortho{} has $\dprobe = \dmodel$ for each model to ensure orthogonality is feasible. Let $L$ be the number of layers in model \model{}. To reduce the influence of a word's semantic properties, we subtract word embeddings from each layer's contextual states
\[
h_i^\ell = h_i^\ell - h_i^0 \hspace{5mm} \forall \ell \in \{1, \ldots, L\}.
\]
For each of \struct{} and \head{}, we train $L$ probes per model---one per layer---and select the probe with the best PTB test metric. For \ortho{}, given limited compute resources and the fact that it is derived from \struct{}, we train a single probe per model on the layer with the best \struct{} PTB test metric. Since the best layer can differ between \struct{} and \head{}, and the objective of \control{} is to control for hidden state quality, we train a separate \control{} instance for each on the corresponding best layer. Control results are reported with the associated probe.

\paragraph{Control Probe Validation.}
We construct a simple evaluation dataset to test how well our control probe abstracts away contextual linguistic information to verify its design. For 500 random words chosen from SimLex999 \citep{hill2014simlex999evaluatingsemanticmodels}, we prompt GPT-4o \cite{openai2024gpt4ocard} to generate 10 sentences that contain the word in substantially different contexts. Refer to Appendix~\ref{sec:control-validation} for the prompt and sample sentences. 

We first measure the variance of each word's \textit{contextual} hidden states (at the layer \control{} is trained at) across the 10 samples, and report the average across all 500 words. We then apply \control{} on these hidden states and recompute this quantity. If the probe successfully abstracts away contextually-determined linguistic information like syntax and retains non-contextual information like lexical semantics, we expect the mean variance of a given word hidden state to significantly decrease, since lexical semantics of a given word should remain the same regardless of context. 

\paragraph{Mechanisms vs. Outcomes Regression.} Our objective is model-level mechanisms vs. outcomes regression because we want to study how effectively the method of syntactic probing predicts an arbitrary model’s latent syntactic knowledge. We compute regressions separately at two BLiMP granularities: (1) overall (averaging predictor and response values across all paradigms in the full dataset), and (2) each of the 13 phenomena separately with Holm-Bonferroni correction. Holm-Bonferroni is an effective method for adjusting significance tests to account for multiple comparisons (13 in our case).

For each linear regression at a given granularity, the predictor variables are probe scores: UUAS (\struct{}, \ortho{}), UAS (\head{}), or Spearman correlation $\rho_s$ (\control{}). We run regressions separately for each probe family, with one data point per model. Fig~\ref{fig:fig_1_design} illustrates the approach: the selected probe for each model is applied to every $S_{acc}$ sentence at the given granularity to compute an average probe score  (\textit{mechanism}), and each $S_{acc}$ and $S_{unacc}$ feed into minimal pair evaluation (\textit{outcome}). 

We fit simple (eqn~\ref{eq:one-probe}) and multiple (eqn~\ref{eq:two-probes}) OLS regressions for each probe family at each granularity:
\begin{gather}
    y = \beta_0 + \beta_1 x_1 + \epsilon \label{eq:one-probe} \\
    y = \beta_0 + \beta_1 x_1 + \beta_2 x_2 + \epsilon \label{eq:two-probes}
\end{gather}
where $y$ is minimal pairs accuracy (eqn~\ref{eq:blimp}) and $x_1$ is the probe score. For multiple regression, $x_1$ is UUAS or UAS and $x_2$ is $\rho_s$. Comparing $\beta_1$ and its $p$-value for the syntax probe when it is the sole predictor versus when paired with the control reveals the extent to which it predicts grammaticality outcomes. Since the two regression models are nested, we also include log-likelihood ratio tests (LRT) for robust statistical comparisons.


\paragraph{Critical Edge Study in Difficult Paradigms.}
Drilling down to the paradigm level, we select the two paradigms in each of the subject--verb agreement (\textit{distractor agreement relational noun}, \textit{distractor agreement relative clause}) and filler--gap (\textit{wh.\ vs.\ that with gap}, \textit{wh.\ vs.\ that with gap long distance}) phenomena that scored lowest in model accuracy in \citet{warstadt-etal-2020-blimp-benchmark}. Rather than tree UUAS/UAS, we determine if the induced syntactic tree for each $S_{acc}$ contains a \emph{critical edge} corresponding to the phenomenon targeted by the minimal pairs---an \texttt{nsubj} edge for subject--verb agreement, and an \texttt{obj} or \texttt{obl} edge for filler--gap. Appendix~\ref{sec:critical-edges} discusses critical edge criteria in-depth. We hypothesize that the binary outcome of the probe's ability to extract the critical edge on $S_{acc}$ is linked to the minimal pair binary outcome.

\section{Results \& Discussion}
\begin{figure}[!htbp]
\centering
  \includegraphics[width=\columnwidth]{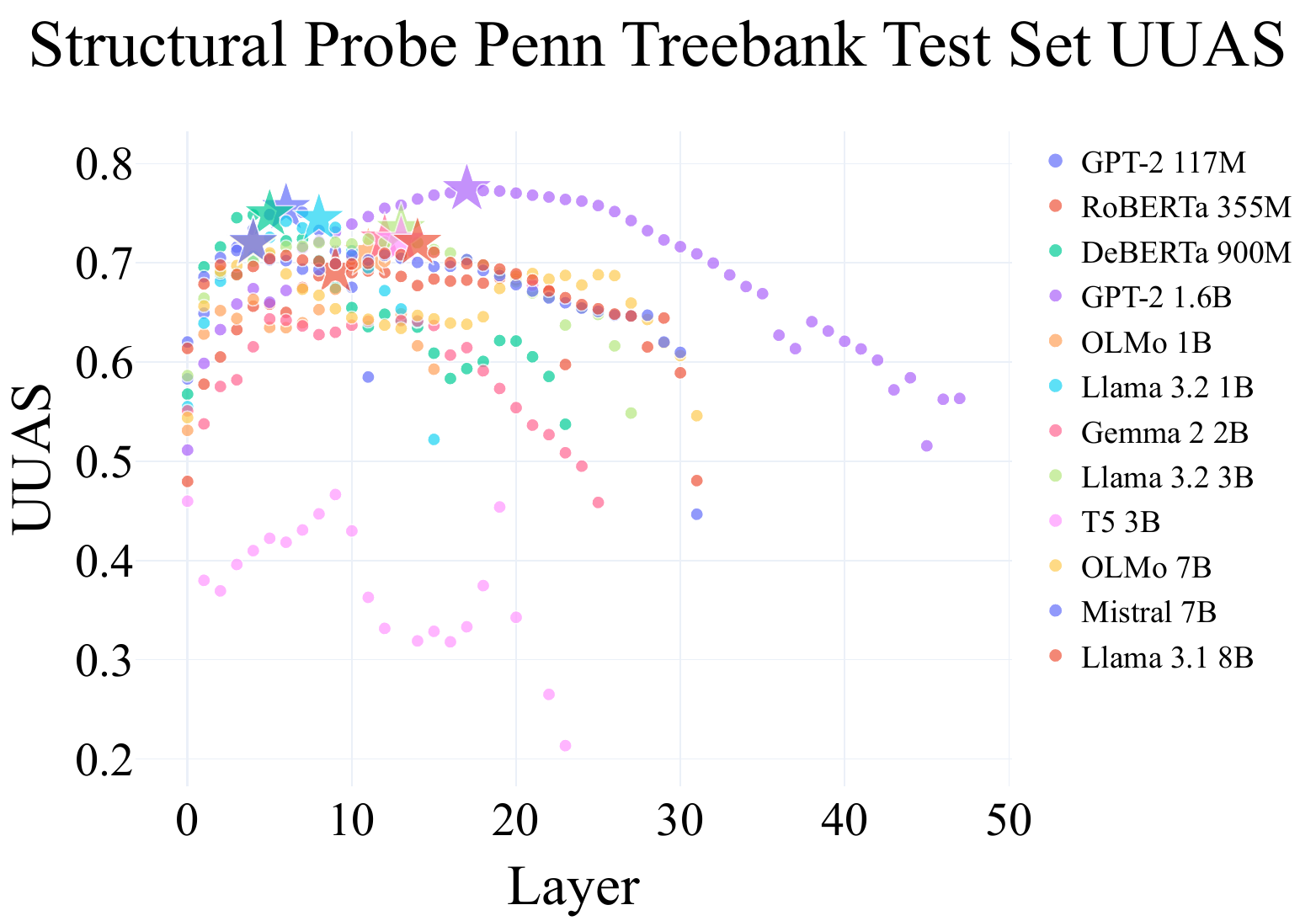}
  \caption{\textbf{Penn Treebank test set \struct{} UUAS for each layer of a sample of our models}. The star icon for a model indicates the layer with the best test set accuracy that is used for BLiMP evaluation.
  For most models, this occurs in the first half. Our results for GPT-2 124M and GPT-2 1.6B closely match those of \citet{eisape2022probingincrementalparsestates}. For T5 3B, although most layers yield low UUAS, the best layer (13) still exceeds 0.7.}
  \label{fig:structural_probe_layers}
\end{figure}

\paragraph{Attachment score typically peaks in the middle layers.} Fig~\ref{fig:structural_probe_layers} plots PTB test \struct{} UUAS across model layers. Best layers for GPT-2 \citep{radford2019language} match with previous results from the literature \citep{eisape2022probingincrementalparsestates}, validating our probe training procedure. In Appendix~\ref{sec:probe-comparison}, Fig~\ref{fig:head_word_probe_layers} plots PTB test \head{} UAS, and we further show that our syntax probes' abilities to recover encoded parse tree information in BLiMP sentences correlate well across training objective (\struct{} vs. \head{}, Fig~\ref{fig:head_struct_comparison}) and architecture (\struct{} vs. \ortho{}, Fig~\ref{fig:ortho_struct_comparison}). Furthermore, the comparable range of UUAS/UAS scores between the PTB test set and BLiMP shows the probes generalize from human-written PTB text to template-generated BLiMP.

\begin{figure}[!htbp]
\centering
    \includegraphics[width=\columnwidth]{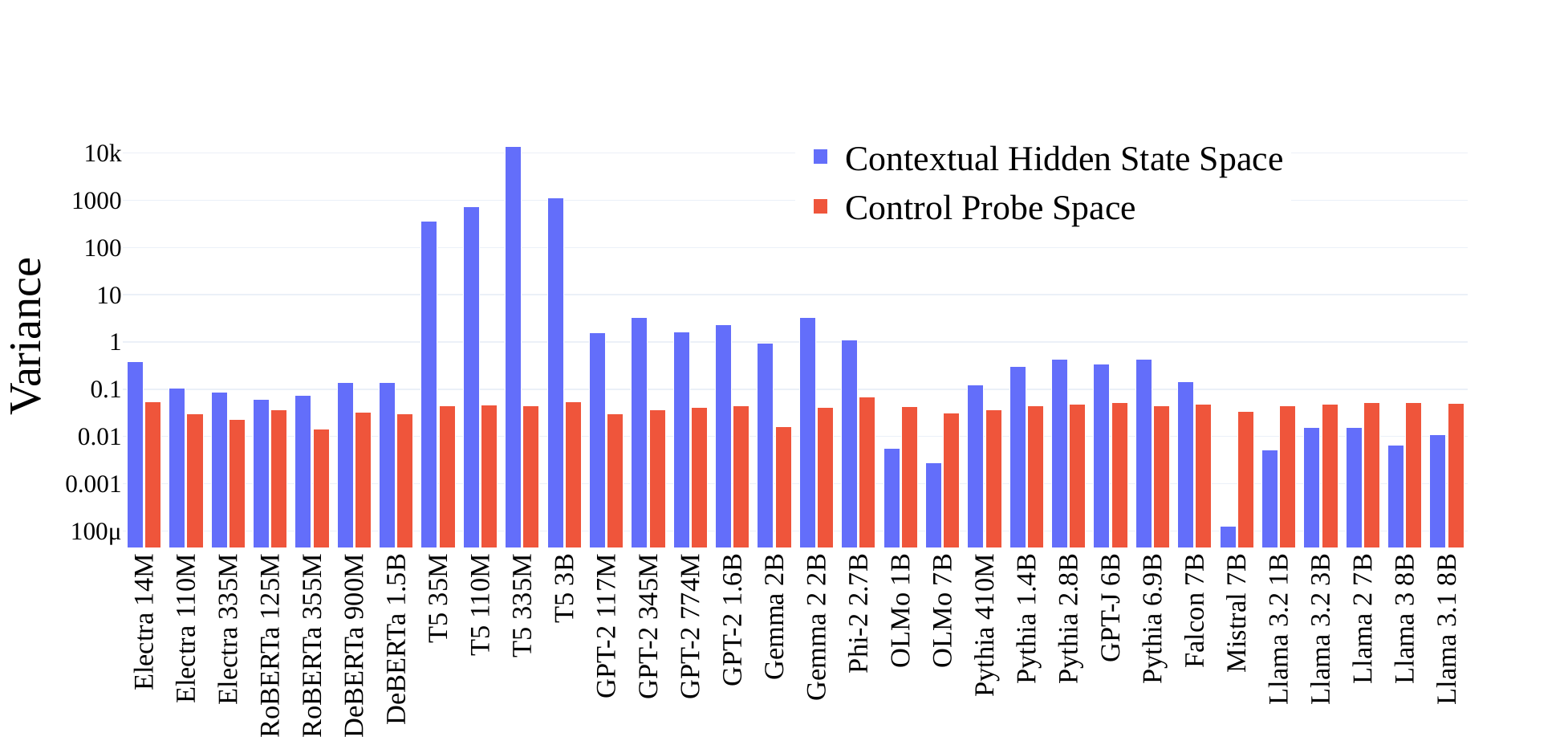}
    \caption{\textbf{Control probes consistently erase contextual information in hidden states, as evidenced by near-zero variance of word contextual hidden states in the projected representation space}. Control probes shown here are trained on structural probe best layers.}
    \label{fig:variance_validation}
\end{figure}

\paragraph{Control probe erases contextual information.}
Fig~\ref{fig:variance_validation} shows that for all models, regardless of the scale of the variance in the average word's contextual hidden states across sentences, the variance in the control probe's projected representation space is near-zero. This shows the probe nullifies context and uncovers \glove{}-like linguistic information within hidden states.

\begin{figure*}[!ht]
\centering
\includegraphics[width=\textwidth]{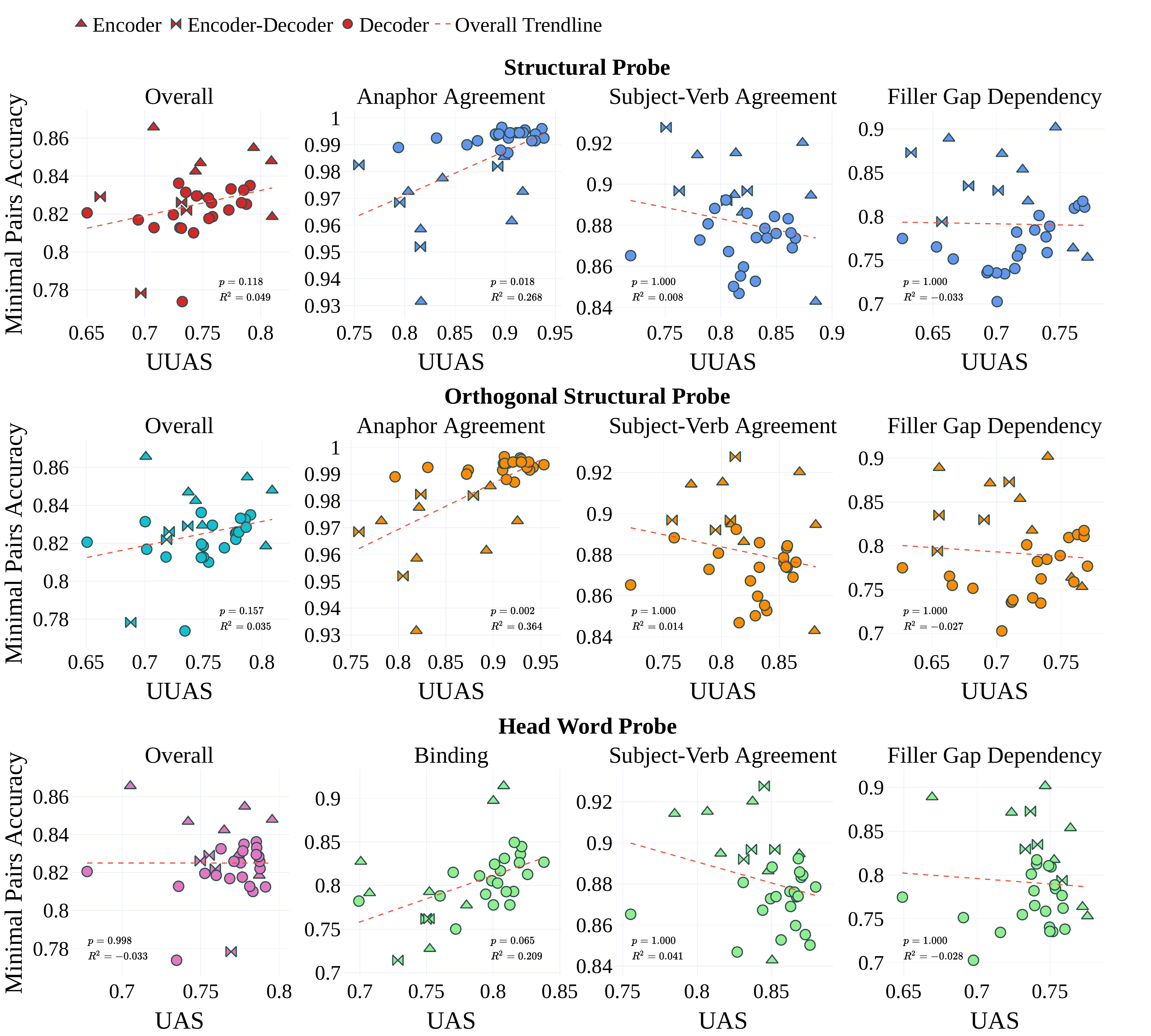}
    \caption{\textbf{Simple regression plots for \struct{} (top row), \ortho{} (middle row), and \head{} (bottom row)}. Each panel is annotated with adjusted $R^2$ and the $p$-value of $\beta_1$. Per-phenomenon results have Holm-Bonferroni correction. The first column of panels shows that at the full dataset granularity, no probe explains the spread in minimal pairs accuracy with any statistical significance. At per-phenomenon granularity, the second column contains the phenomenon with the lowest $p$-value per probe. We additionally highlight subject--verb agreement (third column) and filler--gap (fourth column) as strongly syntactic tasks with critical edges that we identify in Appendix~\ref{sec:critical-edges}.}
    \label{fig:main_results}
\end{figure*}

\begin{table*}[t]
\centering
\begin{subtable}{0.95\textwidth}
\small
\begin{tabular}{cccc|ccc|cc}
\toprule
\multirow{2}{*}{\textbf{Dataset}} & 
\multicolumn{3}{c|}{\textbf{Simple Regression} (\(x_1 = g^{\text{struct}}_{\phi}\))} & 
\multicolumn{3}{c|}{\textbf{Multiple Regression} (\(x_1, x_2 = g^{\text{ctrl}}_{\phi}\))} & 
\multicolumn{2}{c}{\textbf{LRT}} \\
\cmidrule(lr){2-4} \cmidrule(lr){5-7} \cmidrule(lr){8-9}
 & \(\beta_1\) & $p$-value & Adj.\(R^2\) & \(\beta_1\) & $p$-value & Adj.\(R^2\) & Stat & $p$-value \\
\midrule
Overall & 0.133 & 0.117 & 0.049 & 0.109 & 0.214 & 0.051 & 1.161 & 0.281 \\
Anaphor Agreement & 0.165 & \textbf{0.019} & 0.268 & 0.168 & \textbf{0.040} & 0.243 & 0.027 & 1.0 \\
Subject--Verb Agr & -0.108 & 1.0 & 0.007 & -0.047 & 1.0 & 0.128 & 5.236 & 0.243 \\
Filler--Gap Dependency & -0.028 & 1.0 & -0.033 & 0.033 & 1.0 & -0.05 & 0.546 & 1.0 \\
\bottomrule
\end{tabular}
\caption{$x_1 = g_\phi^{\text{struct}}$ UUAS.}
\label{tab:regression_struct_results_main_text}
\end{subtable}

\vspace{0.5em}

\begin{subtable}{0.95\textwidth}
\small
\begin{tabular}{cccc|ccc|cc}
\toprule
\multirow{2}{*}{\textbf{Dataset}} & 
\multicolumn{3}{c|}{\textbf{Simple Regression} (\(x_1 = g^{\text{ortho}}_{\phi}\))} & 
\multicolumn{3}{c|}{\textbf{Multiple Regression} (\(x_1, x_2 = g^{\text{ctrl}}_{\phi}\))} & 
\multicolumn{2}{c}{\textbf{LRT}} \\
\cmidrule(lr){2-4} \cmidrule(lr){5-7} \cmidrule(lr){8-9}
 & \(\beta_1\) & $p$-value & Adj.\(R^2\) & \(\beta_1\) & $p$-value & Adj.\(R^2\) & Stat & $p$-value \\
\midrule
Overall & 0.127 & 0.157 & 0.035 & 0.102 & 0.266 & 0.041 & 1.299 & 0.254 \\
Anaphor Agreement & 0.173 & \textbf{0.002} & 0.364 & 0.173 & \textbf{0.005} & 0.342 & 0.004 & 1.0 \\
Subject--Verb Agr & -0.119 & 1.0 & 0.014 & -0.059 & 1.0 & 0.132 & 5.183 & 0.251 \\
Filler--Gap Dependency & -0.1 & 1.0 & -0.027 & -0.058 & 1.0 & -0.049 & 0.402 & 1.0 \\
\bottomrule
\end{tabular}
\caption{$x_1 = g_\phi^{\text{ortho}}$ UUAS.}
\label{tab:regression_ortho_results_main_text}
\end{subtable}

\vspace{0.5em}

\begin{subtable}{0.95\textwidth}
\small
\begin{tabular}{cccc|ccc|cc}
\toprule
\multirow{2}{*}{\textbf{Dataset}} & 
\multicolumn{3}{c|}{\textbf{Simple Regression} (\(x_1 = g^{\text{head}}_{\phi}\))} & 
\multicolumn{3}{c|}{\textbf{Multiple Regression} (\(x_1, x_2 = g^{\text{ctrl}}_{\phi}\))} & 
\multicolumn{2}{c}{\textbf{LRT}} \\
\cmidrule(lr){2-4} \cmidrule(lr){5-7} \cmidrule(lr){8-9}
 & \(\beta_1\) & $p$-value & Adj.\(R^2\) & \(\beta_1\) & $p$-value & Adj.\(R^2\) & Stat & $p$-value \\
\midrule
Overall & 0.0 & 0.998 & -0.033 & -0.048 & 0.700 & 0.091 & 5.194 & \textbf{0.023} \\
Binding & 0.534 & 0.064 & 0.209 & 0.522 & 0.108 & 0.184 & 0.085 & 1.0 \\
Subject--Verb Agr & -0.205 & 1.0 & 0.041 & -0.221 & 1.0 & 0.025 & 0.56 & 1.0 \\
Filler--Gap Dependency & -0.125 & 1.0 & -0.028 & -0.15 & 1.0 & -0.055 & 0.256 & 1.0 \\
\bottomrule
\end{tabular}
\caption{$x_1 = $ \head{} UAS.}
\label{tab:regression_head_results_main_text}
\end{subtable}

\caption{\textbf{Comparison of simple and multiple regression statistics}. Each subtable corresponds to a different syntax probe as $x_1$, and $x_2 = $ \control{} $\rho_s$. Per-phenomenon $p$-values are Holm-Bonferroni corrected and bold text indicates statistical significance ($p$ < 0.05). The datasets shown correspond to those in Fig~\ref{fig:main_results}. For both \struct{} and \ortho{}, the simple regression model fits anaphor agreement better than the multiple regression model does, but no consistent signal emerges for other phenomena. The higher capacity of \ortho{} relative to \struct{} does not translate to improved regression fit. For \head{}, despite the statistical significance of the LRT at the full dataset granularity, the adjusted $R^2$ of the multiple regression model is very weak. Table~\ref{tab:regression-comparison-full} in Appendix~\ref{sec:full-tables} has regression results for all phenomena.}
\label{tab:regression-comparison}
\end{table*}

\paragraph{No significant fit on overall BLiMP.} The first column of Fig~\ref{fig:main_results} shows OLS simple regression result plots at the full BLiMP dataset granularity. Each of the three syntax probes demonstrates a minimal effect size with none reaching statistical significance. \head{} has no measurable effect at all. Its UAS values generally exceed the UUAS values of other probes, reflecting successful learning of relation directionality, but lack of y-axis stratification prevents predictive power.

Table~\ref{tab:regression-comparison} compares simple and multiple regression statistics. For \head{}, the significant $p$-value of the LRT conveys that considering non-syntactic signal from activations provides better fit on full BLiMP than relation direction-oriented signal alone. Even so, the adjusted $R^2$ is still negligible, showing that neither regression meaningfully explains BLiMP accuracy. For \struct{} and \ortho{}, adding the control probe offers no meaningful improvement.  

\paragraph{Only anaphor agreement has statistical significance at the per-phenomenon granularity.} Out of the three syntax probes and \textit{all} phenomena, the only statistically significant correlations are \struct{} and \ortho{} on anaphor agreement. Strikingly, probes show no predictive power even for phenomena like subject--verb agreement and filler--gap dependencies whose behavior we might expect to be reliant on syntactic representations. Refer to the second, third, and fourth columns of panels in Fig~\ref{fig:main_results} for simple regression plots for these select phenomena.

Table~\ref{tab:regression-comparison} shows that aside from anaphor agreement, there is no consistent result of the simpler regression model providing a statistically significant better fit than the regression that includes the control. As a further example, \head{} encodes hierarchical dominance of the head word. We would expect this to influence Negative Polarity Item (NPI) licensing, as NPIs must be in the structural scope of their licensor. However, our results do not support this expectation; Table~\ref{tab:regression_head_results} reports a corrected $p$-value of 1.0 and an adjusted $R^2$ of zero.

\begin{figure}
\centering
\includegraphics[width=0.95\columnwidth]{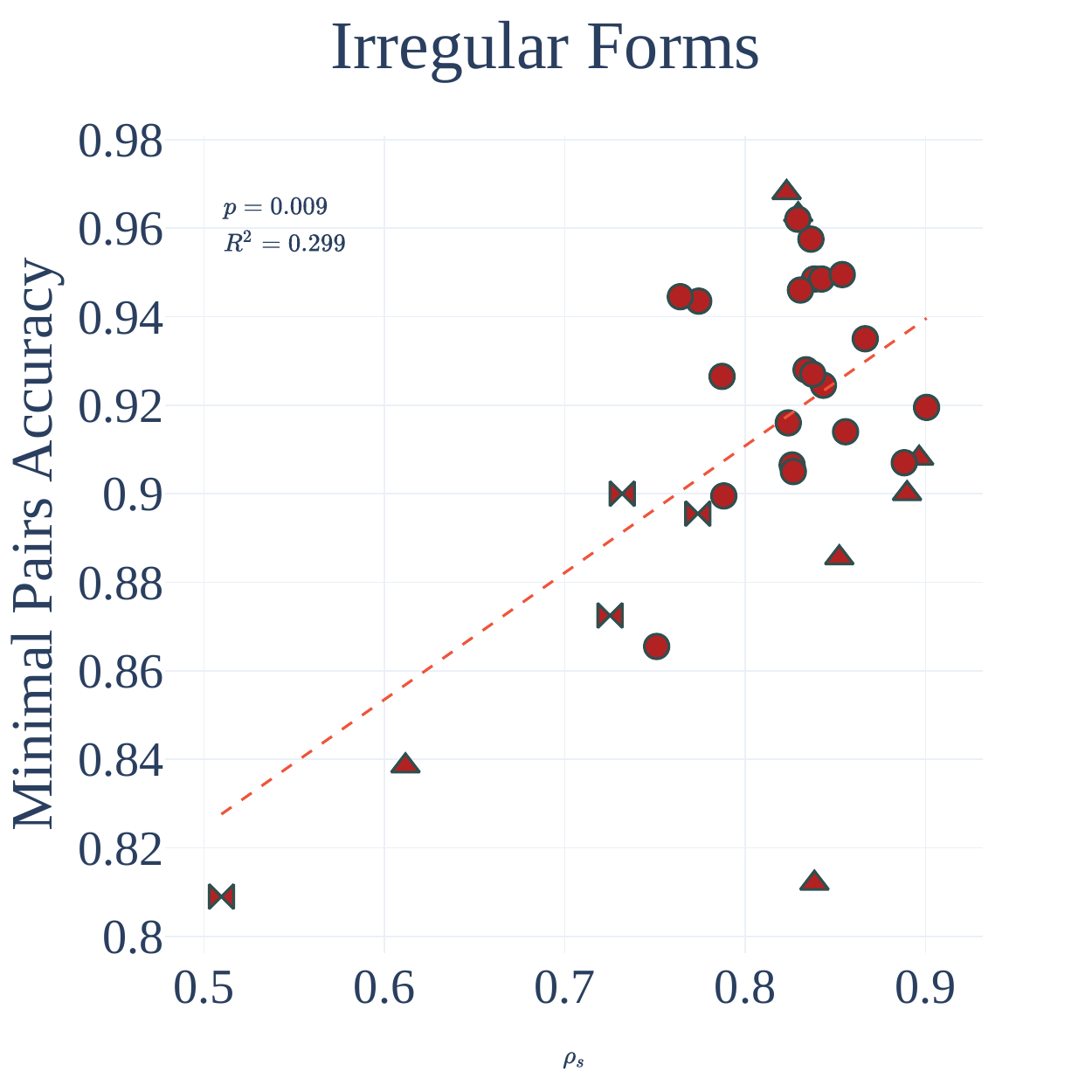}
    \caption{\textbf{Control \control{} (trained on \struct{} best layers) unexpectedly achieves statistical significance for predicting the BLiMP irregular forms phenomenon, which is a morphosyntactic task}. Tables~\ref{tab:simple_ctrl_regression} and \ref{tab:simple_ctrl_regression_head} in Appendix~\ref{sec:full-tables} have full control simple regression results.}
    \label{fig:irregular_forms_control}
\end{figure}

The anaphor agreement result warrants caution. Fig~\ref{fig:irregular_forms_control} shows the surprising result that the control probe, which is trained on a non-morphosyntactic task, can be highly correlated to the BLiMP irregular forms morphosyntactic task. Non-syntactic probes \textit{can} predict syntactic benchmark performance. Without broad systematic correlations, isolated syntax probe--task fits are not evidence that the syntactic information in hidden states that is recoverable by these probes is a primary explanatory variable for downstream behavior; even probes with no syntactic basis can one-off align with syntactic tasks. The anaphor agreement finding could be an artifact of minimal pair accuracy saturation.

\begin{figure*}[t]
    \includegraphics[width=0.95\textwidth]{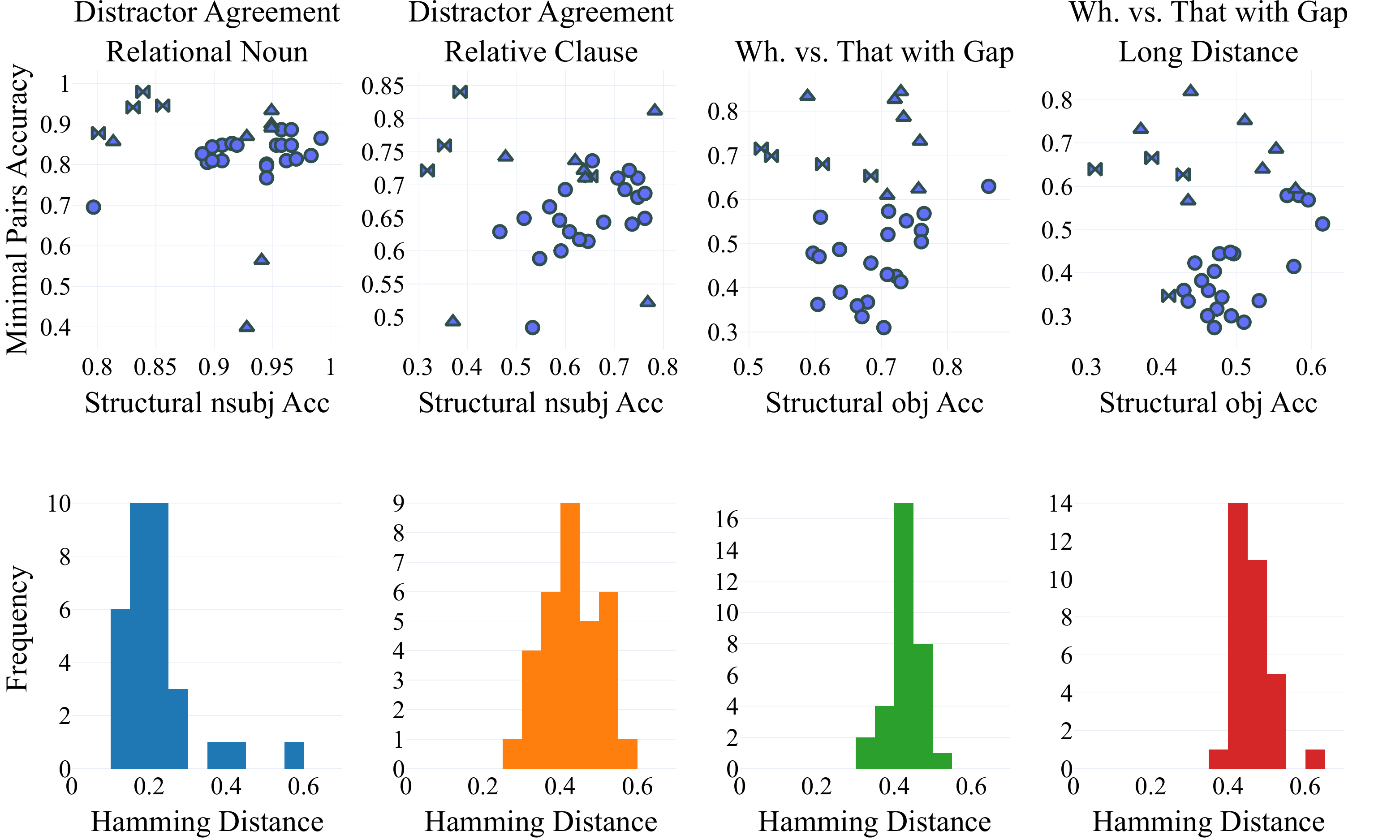}
    \caption{\textbf{Challenging subject--verb agreement and filler--gap phenomena critical edge prediction results for \struct{}}. We compute the Hamming distance between the probe's binary outcome of predicting the critical edge in $S_{acc}$ correctly and the model's binary outcome of resolving the minimal pair correctly. Successful critical edge prediction does not translate to successful minimal pair resolution for 3 out of 4 paradigms. The majority of models show mismatches occurring in approximately 40\% to 60\% of cases (Hamming distance buckets 0.4 to 0.6). Harder paradigms, indicated by lower minimal pair accuracies in the scatter plots, have higher Hamming distance.}
\label{fig:struct_hamming_details}
\end{figure*}

\paragraph{Critical edge predication does not align with minimal pair evaluation.} Fig~\ref{fig:struct_hamming_details} shows that for 3 out of the 4 challenging subject--verb agreement and filler--gap paradigms, the match rate between \struct{} correctly predicting the critical edge and the model resolving the minimal pair correctly is between 40--60\% for the majority of models. The wide spread of minimal pair accuracies along the y-axis of the scatter plots underscores the relative difficulty of these paradigms; by contrast, in Fig~\ref{fig:main_results} plots most accuracies cluster well above 0.7. However, high Hamming distances between edge prediction and minimal pair evaluation mean that errors do not align as much as expected. Extracting critical edges, despite relevance to deciding between the words that are different between the minimal pairs (Appendix~\ref{sec:critical-edges}), does not in fact predict minimal pair outcome. Fig~\ref{fig:head_hamming_details} in Appendix~\ref{sec:full-hamming} yields the same conclusion from \head{} results.

\begin{figure}
\centering
\includegraphics[width=\columnwidth]{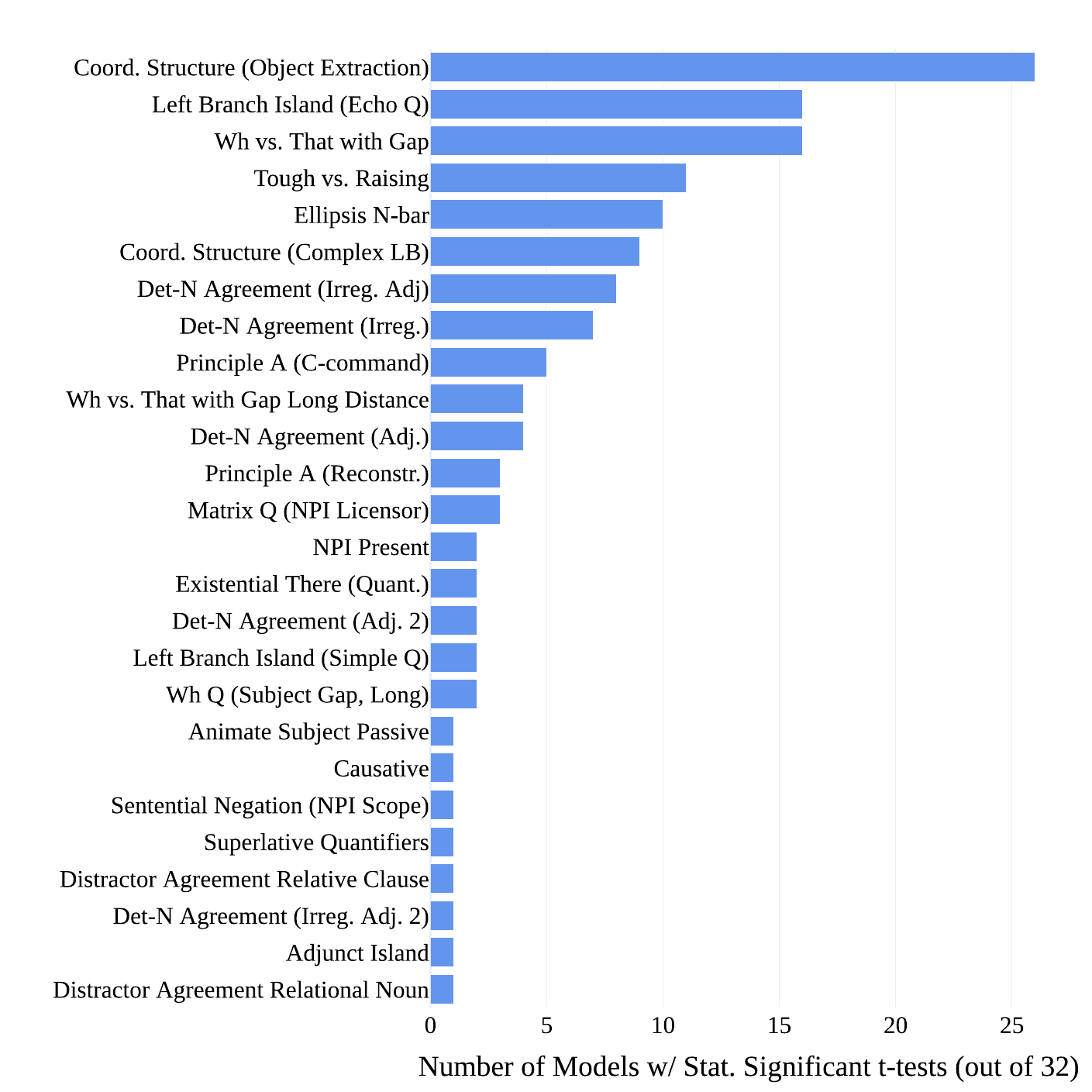}
    \caption{\textbf{Sentence-level UUAS score distribution $t$-test results}. Out of 67 BLiMP paradigms, only the 26 shown here have at least one model that has a statistically significant difference in the UUAS score means.}
    \label{fig:sentence_level}
\end{figure}

\paragraph{UUAS score distributions largely overlap for minimal pair outcomes.} While cross-model correlations between probing and downstream performance have proved elusive, a natural follow-up investigation for each model is the sentence-level comparison of if there is a statistically significant difference in the UUAS score distribution between the pool of sentences where the minimal pair is evaluated correctly and the pool where it is evaluated incorrectly. For each BLiMP paradigm, we run a two-sample one-sided $t$-test for each model on these pools of UUAS scores and apply Holm-Bonferroni correction on the test $p$-values. The results in Fig~\ref{fig:sentence_level} show that only 8 out of 67 BLiMP paradigms have statistical evidence for the mean of UUAS scores for correctly predicted minimal pairs being higher than that of incorrectly predicted minimal pairs for at least 5 models. Yet again, we conclude that for most models and paradigms, UUAS score of the probe at the sentence level (mechanism) does not predict whether the model will succeed at the corresponding minimal pair evaluation (outcome).

\section{Related Work}
Our work builds on the extensive literature examining the syntactic capabilities of language models through probing \citep[among others]{hewitt-manning-2019-structural, clark2019bert, muller2022probing, Limisiewicz2021orth, diego2024polar}.

\paragraph{Causal Analysis.} Recent probing work aims to make \textit{causal} claims between probed representations and observed model downstream behavior. Causal analysis intervenes in model representations to understand their effects. \citet{tucker2021modified} demonstrate that counterfactual interventions on contextual embeddings—designed to erase information identified by a probe—can predictably affect downstream behavior. However, they find that such effects are not robust across models or probes. Similarly, \citet{eisape2022probingincrementalparsestates} perform causal intervention on transformer states during autoregressive generation from GPT-2. \citet{arora-etal-2024-causalgym} introduce a causal interpretability method benchmark, CausalGym, to evaluate interventions sourced from minimal pairs on next-token predictions. Rather than evaluating causality, we focus on probing's statistical predictive power.

\paragraph{Advancing Probing Research.} \citet{probing-promises-shortcomings} surveys the promises and shortcomings of probing research. Our paper addresses several core issues raised and advances the discussion with new empirical evidence. First, Belinkov emphasizes the importance of controls for confounders. We introduce a control probe trained to recover non-syntactic information and demonstrate that it can show correlations with syntactic benchmarks. Second, he notes probe complexity affects interpretation, with simpler probes (e.g., linear) preferred to avoid the probe independently learning the probing task. Our work uses three linear probes with different objective functions and metrics for robustness. Finally, Belinkov warns that probing experiments may conflate properties of the model with properties of the datasets used for training or probing, making it difficult to disentangle the two. We use different datasets for training probes (Penn Treebank) and evaluation (BLiMP), which helps ensure our findings are not artifacts of a single dataset.

While some shortcomings of probing are already known, systematic studies like ours are necessary to thoroughly test hypotheses. Notably, our scope spans a broader range of models and linguistic properties than prior work, including \citet{ravichander2021probingprobingparadigmdoes}, who focus on verb tense, subject number, and object number in Natural Language Inference with pre-transformer architectures, and \citet{elazar2021amnesicprobingbehavioralexplanation}, who apply amnesic probing to sequence tagging tasks with BERT \cite{devlin2019bertpretrainingdeepbidirectional}. 



\paragraph{Future Multilingual Study.}
Just within the last few months, we have seen new work substantially advancing BLiMP for multilingual use cases. The largest cross-lingual development is MultiBLiMP 1.0 \cite{jumelet2025multiblimp10massivelymultilingual}, which has over 128,000 minimal pairs across 101 languages specifically for subject--verb agreement. Furthermore, \citet{başar2025turblimpturkishbenchmarklinguistic} contribute TurBLiMP, a Turkish minimal pairs benchmark with 16 linguistic phenomena. We emphasize in our work that a model's ability to solve minimal pairs correctly is a desired linguistic property and critical evaluation tool, which \citet{başar2025turblimpturkishbenchmarklinguistic} also echo. Much of our evaluation framework is applicable to these new benchmarks: MultiBLiMP uses Universal Dependencies and TurBLiMP includes BLiMP phenomena. As momentum builds towards targeted syntactic analysis for more and lower-resource languages, we believe that future work on our framework towards multilinguality would help identify language-specific probing and downstream performance artifacts. While probing has little predictive power in English, replicating the study could yield novel insights from languages with for instance, richer subject--verb agreement systems than English or greater word order flexibility as seen in Turkish.

\section{Conclusion}
Targeted evaluations like BLiMP are widely used to benchmark model syntactic knowledge. Given the importance of minimal pairs, our key contribution is to show systematically that probing fails to reliably extract the latent syntactic knowledge within models that correlates with minimal pair accuracy. The high BLiMP scores observed across the open-weight models studied suggest effective learning of syntactic phenomena, and probing task performance is likewise strong. However, across models, probing performance does not align with BLiMP performance, even in fine-grained settings where such alignment might reasonably be expected. 

Returning to our ``mechanisms vs. outcomes'' framework, the mechanism of recovering syntactic knowledge from model activations through linear probing \emph{does not} at all predict targeted outcomes. A common question in probing research is whether the probed representations are actually used to produce model outputs, motivating the line of work on causality. However, our negative finding regarding predictive power preempts this causal question. Extensive research on probing and interpretability more broadly has advanced our knowledge, but a robust understanding of how syntax is represented within models remains to be developed.

\section{Limitations}

Our evaluation only uses English grammatical benchmarks and could usefully be extended to other languages. Extending our work to multilingual benchmarks would strengthen the study. This would require formalization of a unified evaluation framework to handle differing linguistic phenomena across languages. Moreover, while we design our syntax probes to be lightweight and to learn to extract syntax with different training objectives, they do not encompass all possible syntactic probing tasks or architectural variants. Other probe architectures could yield different results. Due to compute limitations, we also do not train probes on models larger than 8B parameters.



\section*{Acknowledgements}
The authors would like to thank the anonymous reviewers for constructive feedback during the review period. CM is a fellow in the CIFAR Learning in Machines and
Brains program. We thank members of the Stanford NLP Group for their insightful comments and support throughout the project. 

\bibliography{custom}

\appendix
\clearpage
\onecolumn
\section{Models}\label{sec:full-model-list}
\begin{table}[!htbp]
\centering
\resizebox{!}{!}{
\begin{tabular}{@{}ll@{}}
\toprule
Model                         & HuggingFace ID                     \\ \midrule
\emph{Architecture: Decoder}         &                                    \\
GPT-2 124M \cite{radford2019language}                    & \texttt{gpt2}                               \\
GPT-2 345M                     & \texttt{gpt2-medium}                        \\
GPT-2 774M                     & \texttt{gpt2-large}                         \\
GPT-2 1.6B                     & \texttt{gpt2-xl}                            \\
Pythia 410M \cite{biderman2023pythia}                   & \texttt{EleutherAI/pythia-410m}             \\
Pythia 1.4B                   & \texttt{EleutherAI/pythia-1.4b}             \\
Pythia 2.8B                   & \texttt{EleutherAI/pythia-2.8b}             \\
Pythia 6.9B                   & \texttt{EleutherAI/pythia-6.9b}             \\
GPT-J 6B \cite{gpt-j}                      & \texttt{EleutherAI/gpt-j-6b}                \\
Falcon 7B \cite{almazrouei2023falconseriesopenlanguage}                     & \texttt{tiiuae/falcon-7b}                   \\
Llama 3.2 1B \cite{grattafiori2024llama3herdmodels}                 & \texttt{meta-llama/Llama-3.2-1B}            \\
Llama 3.2 3B                  & \texttt{meta-llama/Llama-3.2-3B}            \\
Llama 2 7B \cite{touvron2023llama2openfoundation}                    & \texttt{meta-llama/Llama-2-7b-hf}           \\
Llama 3 8B                    & \texttt{meta-llama/Meta-Llama-3-8B}         \\
Llama 3.1 8B                  & \texttt{meta-llama/Meta-Llama-3.1-8B}       \\
Mistral 7B \cite{jiang2023mistral7b}                   & \texttt{mistralai/Mistral-7B-v0.1}          \\
Gemma 2B \cite{gemmateam2024gemmaopenmodelsbased}                      & \texttt{google/gemma-2b}                    \\
Gemma 2 2B \cite{gemmateam2024gemma2improvingopen}                    & \texttt{google/gemma-2-2b}                  \\
OLMo 1B \cite{Groeneveld2023OLMo}                      & \texttt{allenai/OLMo-1B-hf}                 \\
OLMo 7B \cite{Groeneveld2023OLMo}                       & \texttt{allenai/OLMo-1.7-7B-hf}             \\
Phi 2 2.7B \cite{gunasekar2023textbooksneed}                   & \texttt{microsoft/phi-2}                    \\ \midrule
\emph{Architecture: Encoder}         &                                    \\
ELECTRA-Small 14M  \cite{clark2020electrapretrainingtextencoders}                 & \texttt{google/electra-small-discriminator} \\
ELECTRA-Base 110M                  & \texttt{google/electra-base-discriminator}  \\
ELECTRA-Large 335M                 & \texttt{google/electra-large-discriminator} \\
RoBERTa 125M \cite{DBLP:journals/corr/abs-1907-11692}                 & \texttt{FacebookAI/roberta-base}            \\
RoBERTa 355M                  & \texttt{FacebookAI/roberta-large}           \\
DeBERTa v2 900M \cite{he2021deberta}              & \texttt{microsoft/deberta-v2-xlarge}        \\
DeBERTa v2 1.5B               & \texttt{microsoft/deberta-v2-xxlarge}       \\ \midrule
\emph{Architecture: Encoder-Decoder} &                                    \\
T5 35M \cite{2020t5}                        & \texttt{google-t5/t5-small}                 \\
T5 110M                       & \texttt{google-t5/t5-base}                  \\
T5 335M                       & \texttt{google-t5/t5-large}                 \\
T5 3B                         & \texttt{google-t5/t5-3b}\\
\bottomrule
\end{tabular}}
\caption{All 32 models used for every experiment.}
\end{table}
\twocolumn
\section{Probe Training Details}
\label{sec:probe-training}
All training and inference occurs on a single NVIDIA RTX 6000 Ada Generation or NVIDIA RTX A6000 GPU depending on cluster availability. Training probes on the largest models could take up to 4 days to complete. In the Penn Treebank, the training set has 39,831 sentences, validation set 1,700 sentences, and test set 2,416 sentences. In the following sections, we specify training hyperparameters for each probe.
\subsection{Structural Probe (\struct{})}
\begin{itemize}
    \item Batch size: 32
    \item Max epochs: 300
    \item Early stopping criteria: validation loss; patience of 50 epochs; eval interval of 1 epoch
    \item Learning rate: $1e^{-4}$; AdamW optimizer
    \item Learning rate schedule: Linear decay with warmup for 10\% of max epochs
\end{itemize}

\subsection{Orthogonal Structural Probe (\ortho{})}
\begin{itemize}
    \item Batch size: 32, 64
    \item Max epochs: 50
    \item Early stopping criteria: validation loss; patience of 5 epochs; eval interval of 1 epoch
    \item Learning rate: $1e^{-4}$; AdamW optimizer
\end{itemize}

\subsection{Head Word Probe (\head{})}
\begin{itemize}
    \item Batch size: 32
    \item Max epochs: 300
    \item Early stopping criteria: validation UAS; patience of 50 epochs; eval interval of 1 epoch
    \item Learning rate: $1e^{-4}$; AdamW optimizer
    \item Learning rate schedule: Linear decay with warmup for 10\% of max epochs
\end{itemize}

\subsection{Control Probe (\control{})}
\begin{itemize}
    \item Batch size: 128
    \item Max epochs: 300
    \item Early stopping criteria: validation loss; patience of 50 epochs; eval interval of 1 epoch
    \item Learning rate: $1e^{-4}$; AdamW optimizer
    \item Learning rate schedule: Linear decay with warmup for 10\% of max epochs
\end{itemize}

\section{Control Probe Validation}
\label{sec:control-validation}
We prompt GPT-4o to elicit sentences with varied contextual usage for a given word:

\begin{quote}
I will give you a word. Generate 10 sentences that use the word in substantially different contexts. Do not capitalize or add any punctuation to the word, but you can add punctuation elsewhere in the sentences. Do not make the word the first word of the sentence; instead it should be near the end! 
Return only the list of sentences in the format of a Python list without Markdown. 
Word: \{\{word\}\}    
\end{quote}

\noindent Outputs for the words ``window'' and ``flour'':

\begin{lstlisting}
"window": [
    "The cat loves to sit by the window and watch the birds.",
    "After the update, the software has a new window that is more user-friendly.",
    "She opened the window to let in some fresh air.",
    "There was a brief window of opportunity to complete the task.",
    "The display showed a colorful window of stained glass.",
    "He peered through the window to see what was happening inside.",
    "The startup process will initiate a pop-up window for further instructions.",
    "In the story, the hero jumps out of a high window to escape.",
    "The interview was scheduled within a tight window of time.",
    "She painted a beautiful scene of a sunrise in the window of her art studio."
],
"flour": [
    "To bake the cake, you'll need to sift the flour thoroughly.",
    "She dusted the countertop with flour before kneading the dough.",
    "The recipe calls for almond flour instead of regular wheat.",
    "A fine layer of flour covered the baker's apron after a long day.",
    "If you want a crispy crust, coat the chicken with flour before frying.",
    "They decided to try a gluten-free flour for their cookies this time.",
    "The bread's texture depends largely on the quality of the flour used.",
    "After spilling the bag, a cloud of flour filled the air in the kitchen.",
    "He couldn't believe how much more expensive organic flour was at the store.",
    "The children made a mess while trying to measure the flour for the pancakes."
]
\end{lstlisting}

\section{Critical Tree Edges for BLiMP Phenomena} \label{sec:critical-edges}
We investigate if the syntax probe's accuracy in predicting a \textit{critical edge} we define on a per-paradigm basis in the acceptable sentence's dependency parse tree is consistent with the model's outcome of assigning the acceptable sentence a higher log probability than the unacceptable one. The critical edge involves the minimal difference in the pair's sentences. In the following sections, we justify the critical edges for the subject--verb agreement and filler--gap paradigms that we experiment with:
\begin{itemize}
    \item Subject--verb agreement
        \begin{itemize}
            \item Distractor agreement relational noun
            \item Distractor agreement relative clause
        \end{itemize}
    \item Filler--gap dependencies
    \begin{itemize}
        \item \textit{Wh.}\ vs.\ \textit{that} with gap
        \item \textit{Wh.}\ vs.\ \textit{that} with gap long distance
    \end{itemize}
\end{itemize}
\subsection{Subject--Verb Agreement}
\begin{figure*}
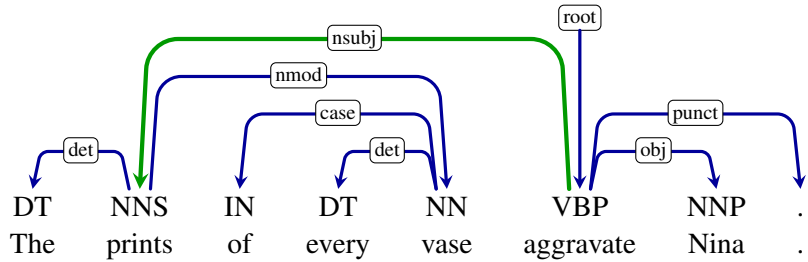

\centering
\begin{dependency}[edge style={blue!60!black,very thick}]
\begin{deptext}[column sep=.5cm, row sep=.1ex]
DT \& NNS \& IN \& DT \& NN \& VBP \& NNP \& . \\
The \& prints \& of \& every \& vase \& aggravate \& Nina \& .\\
\end{deptext}
\deproot[edge height=2.5cm]{6}{root}
\depedge{2}{1}{det}
\depedge[edge style={green!60!black,ultra thick}]{6}{2}{nsubj}
\depedge{6}{7}{obj}
\depedge{6}{8}{punct}
\depedge{5}{4}{det}
\depedge{5}{3}{case}
\depedge{2}{5}{nmod}
\end{dependency}
\caption{\textbf{\textit{Distractor agreement relational noun} sample well-formed sentence Stanza dependency parse tree}. The \texttt{nsubj} edge in green is the critical edge.}
\label{fig:relational-noun}
\end{figure*}

The critical edge is one that connects the relevant subject and the verb targeted by the minimal pair. Consider the following test instance from the \textit{distractor agreement relational noun} suite:

\begin{flushleft}
    The prints of every vase \underline{aggravate} Nina. \hfill(acc.) \\
    The prints of every vase \underline{aggravates} Nina. \hfill(unacc.)
\end{flushleft}
Agreement on the verb `aggravate' is determined by its subject, the plural noun `prints'. This is represented in parse trees as an edge connecting `prints' and `aggravate,' labeled \texttt{nsubj}. Fig~\ref{fig:relational-noun} shows the parse tree for the acceptable sentence. If a model's syntactic representations are related to its success on subject--verb agreement, \texttt{nsubj} is the critical dependency to evaluate -- a model successfully representing this edge means that it is minimally aware of what noun is its subject, and thus, which noun it should use to determine whether `aggravate' or `aggravates' should be assigned a higher log probability (see \citealp{clark2019bert, jian-reddy-2023-syntactic} for similar discussions of syntactic attention heads).

\begin{figure*}
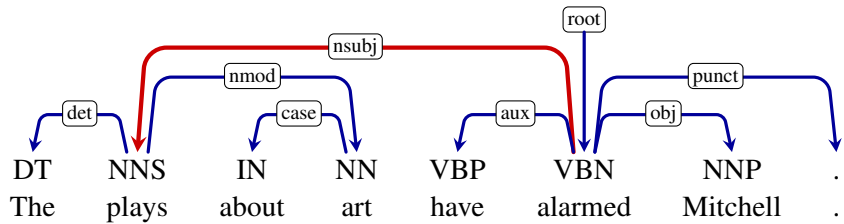

\centering
\begin{dependency}[edge style={blue!60!black,very thick}]
\begin{deptext}[column sep=.5cm, row sep=.1ex]
DT \& NNS \& IN \& NN \& VBP \& VBN \& NNP \& . \\
The \& plays \& about \& art \& have \& alarmed \& Mitchell \& .\\
\end{deptext}
\deproot{6}{root}
\depedge{2}{1}{det}
\depedge[edge unit distance=2ex, edge style={red!80!black,ultra thick}]{6}{2}{nsubj}
\depedge{2}{4}{nmod}
\depedge{4}{3}{case}
\depedge{6}{5}{aux}
\depedge{6}{7}{obj}
\depedge{6}{8}{punct}
\end{dependency}
\caption{\textbf{\textit{Distractor agreement relative clause} sample that we filter out since the critical word is the auxiliary ``have'', which is not connected to the subject ``plays''}.}
\label{fig:subjverb-bad}
\end{figure*}

Under the UD formalism whose parses Stanza produces, the verb which agrees with the subject is not always the head of the \texttt{nsubj} relation in a sentence. For example, in Fig~\ref{fig:subjverb-bad} the verb which agrees with the subject is the auxiliary `have', but the UD parse assigns `alarmed,' which does not agree, as the head. For these critical edge experiments only, we filter out all incongruent cases where UD assigns no edge between the subject and critical verb, which leaves \textbf{236} valid minimal pairs for \textit{distractor agreement relational noun} and \textbf{345} for \textit{distractor agreement relative clause}. 


\begin{figure*}
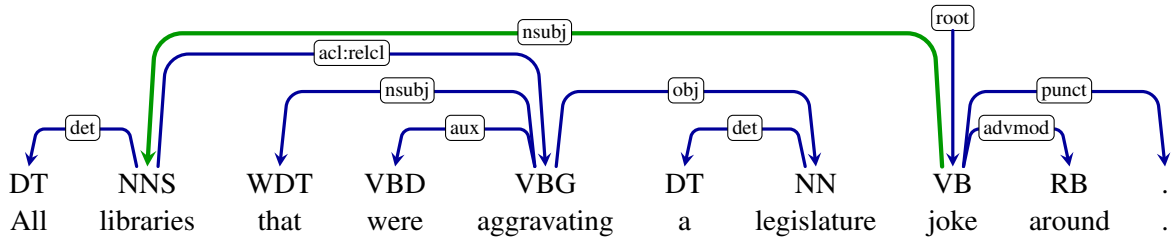

\centering
\begin{dependency}[edge style={blue!60!black,very thick}]
\begin{deptext}[column sep=.5cm, row sep=.1ex]
DT \& NNS \& WDT \& VBD \& VBG \& DT \& NN \& VB \& RB \& . \\
All \& libraries \& that \& were \& aggravating \& a \& legislature \& joke \& around \& .\\
\end{deptext}
\deproot[edge height=2.2cm]{8}{root}
\depedge{2}{1}{det}
\depedge[edge unit distance=1.7ex, edge style={green!60!black,ultra thick}]{8}{2}{nsubj}
\depedge{2}{5}{acl:relcl}
\depedge{5}{3}{nsubj}
\depedge{5}{4}{aux}
\depedge{5}{7}{obj}
\depedge{7}{6}{det}
\depedge{8}{9}{advmod}
\depedge{8}{10}{punct}
\end{dependency}
\caption{\textbf{\textit{Distractor agreement relative clause} sample well-formed sentence Stanza dependency parse tree}. The \texttt{nsubj} edge in green is the critical edge.}
\label{fig:relative-clause}
\end{figure*}

\begin{figure*}[!htbp]
\centering
\begin{dependency}[edge style={blue!60!black,very thick}]
\begin{deptext}[column sep=.5cm, row sep=.1ex]
NNP \& VBD \& VBN \& WP \& DT \& NN \& VBD \& . \\
Marcus \& had \& remembered \& who \& some \& lady \& disliked \& .\\
\end{deptext}
\deproot{3}{root}
\depedge{3}{1}{nsubj}
\depedge[edge unit distance=1.7ex, edge style={green!60!black,ultra thick}]{7}{4}{obj}
\depedge{3}{2}{aux}
\depedge[edge unit distance=1.85ex]{3}{8}{punct}
\depedge[edge unit distance=1.7ex]{3}{7}{ccomp}
\depedge{6}{5}{det}
\depedge{7}{6}{nsubj}
\end{dependency}
\caption{\textbf{\textit{Wh.\ vs.\ that with gap} sample well-formed sentence Stanza dependency parse tree}. The \texttt{obj} edge in green is the critical edge.}
\label{fig:wh-vs-that}
\end{figure*}

\begin{figure*}
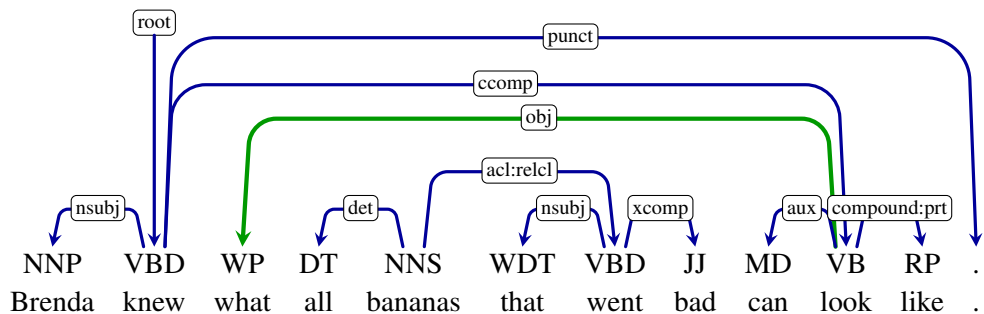

\centering
\begin{dependency}[edge style={blue!60!black,very thick}]
\begin{deptext}[column sep=.2cm, row sep=.1ex]
NNP \& VBD \& WP \& DT \& NNS \& WDT \& VBD \& JJ \& MD \& VB \& RP \& . \\
Brenda \& knew \& what \& all \& bananas \& that \& went \& bad \& can \& look \& like \& .\\
\end{deptext}
\deproot[edge height=3.2cm]{2}{root}
\depedge{2}{1}{nsubj}
\depedge[edge unit distance=1.6ex]{2}{12}{punct}
\depedge[edge unit distance=1.55ex]{2}{10}{ccomp}
\depedge[edge unit distance=1.4ex, edge style={green!60!black,ultra thick}]{10}{3}{obj}
\depedge{5}{4}{det}
\depedge{5}{7}{acl:relcl}
\depedge{7}{6}{nsubj}
\depedge{7}{8}{xcomp}
\depedge{10}{9}{aux}
\depedge{10}{11}{compound:prt}
\end{dependency}
\caption{\textbf{\textit{Wh.\ vs.\ that with gap long distance} sample well-formed sentence Stanza dependency parse tree}. The \texttt{obj} edge in green is the critical edge.}
\label{fig:wh-vs-that-long}
\end{figure*}
\subsection{Filler-gap}
We establish the critical edge for \textit{wh.\ vs.\ that with gap} and \textit{wh.\ vs.\ that with gap long distance} in the same way. Fig~\ref{fig:wh-vs-that} and Fig~\ref{fig:wh-vs-that-long} show sample acceptable parse trees for \textit{wh.\ vs.\ that with gap} and \textit{wh.\ vs.\ that with gap long distance}. The critical edge is the edge between the \textit{wh}-word and the verb it is the direct (\texttt{obj}) or oblique object (\texttt{obl}) of; any other case is filtered out. After filtering, there are \textbf{972} valid minimal pairs for \textit{wh.\ vs.\ that with gap} and \textbf{885} for \textit{wh.\ vs.\ that with gap long distance}. 

\FloatBarrier
\clearpage
\onecolumn

\section{Probe Training and Comparison}
\label{sec:probe-comparison}

\begin{figure}[!htbp]
\centering
  \includegraphics[width=0.85\textwidth]{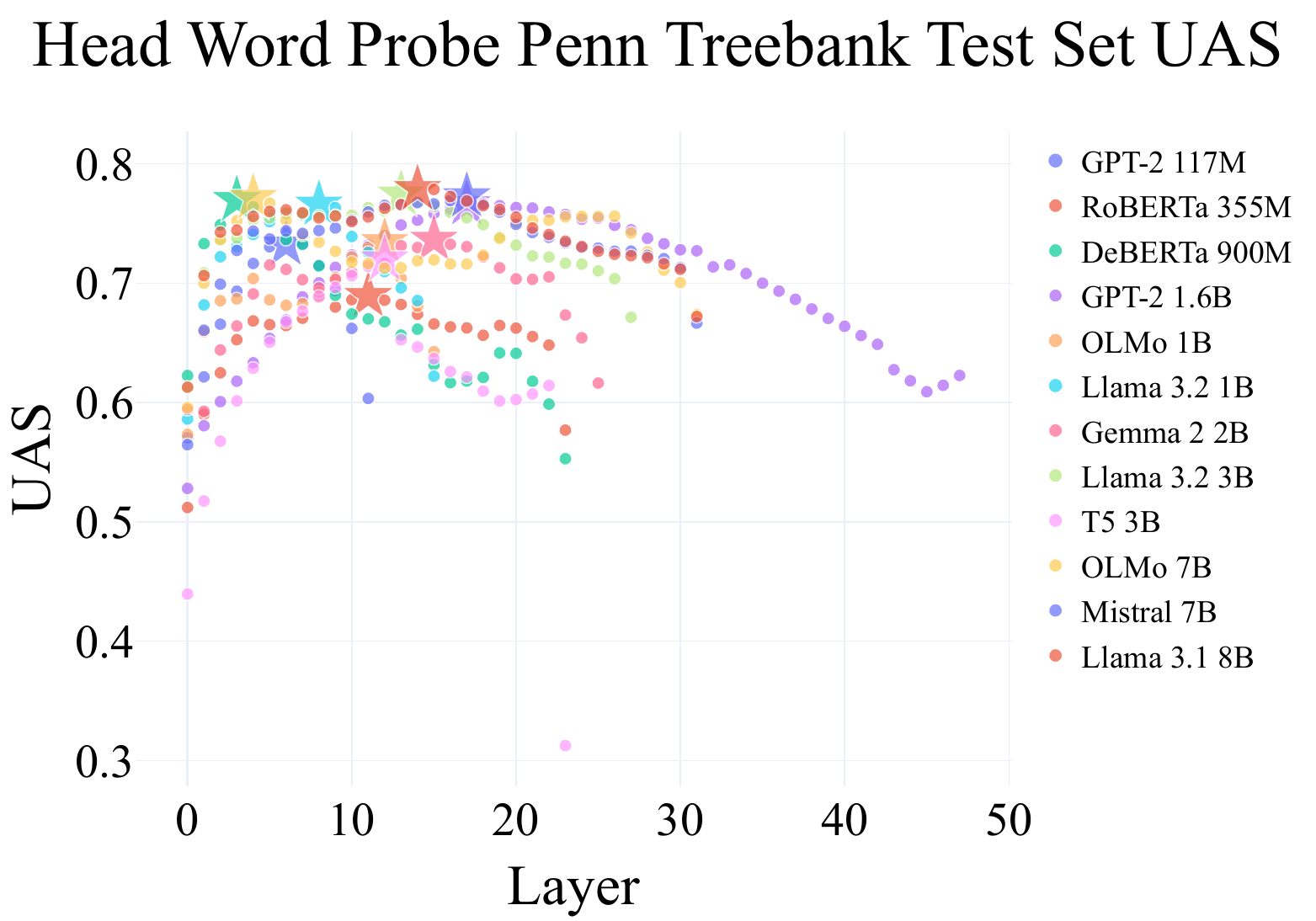}
  \caption{\textbf{Penn Treebank test set \head{} UAS for each layer of a sample of our models}. We train a probe on each layer of the model using the train and validation splits and select the probe for the layer with the best test set accuracy -- indicated in the plot with a star icon -- for BLiMP evaluation. For most models, this occurs in the first half.}
  \label{fig:head_word_probe_layers}
\end{figure}
\FloatBarrier
\clearpage

\begin{figure}[!htbp]
\centering
  \includegraphics[width=0.85\textwidth]{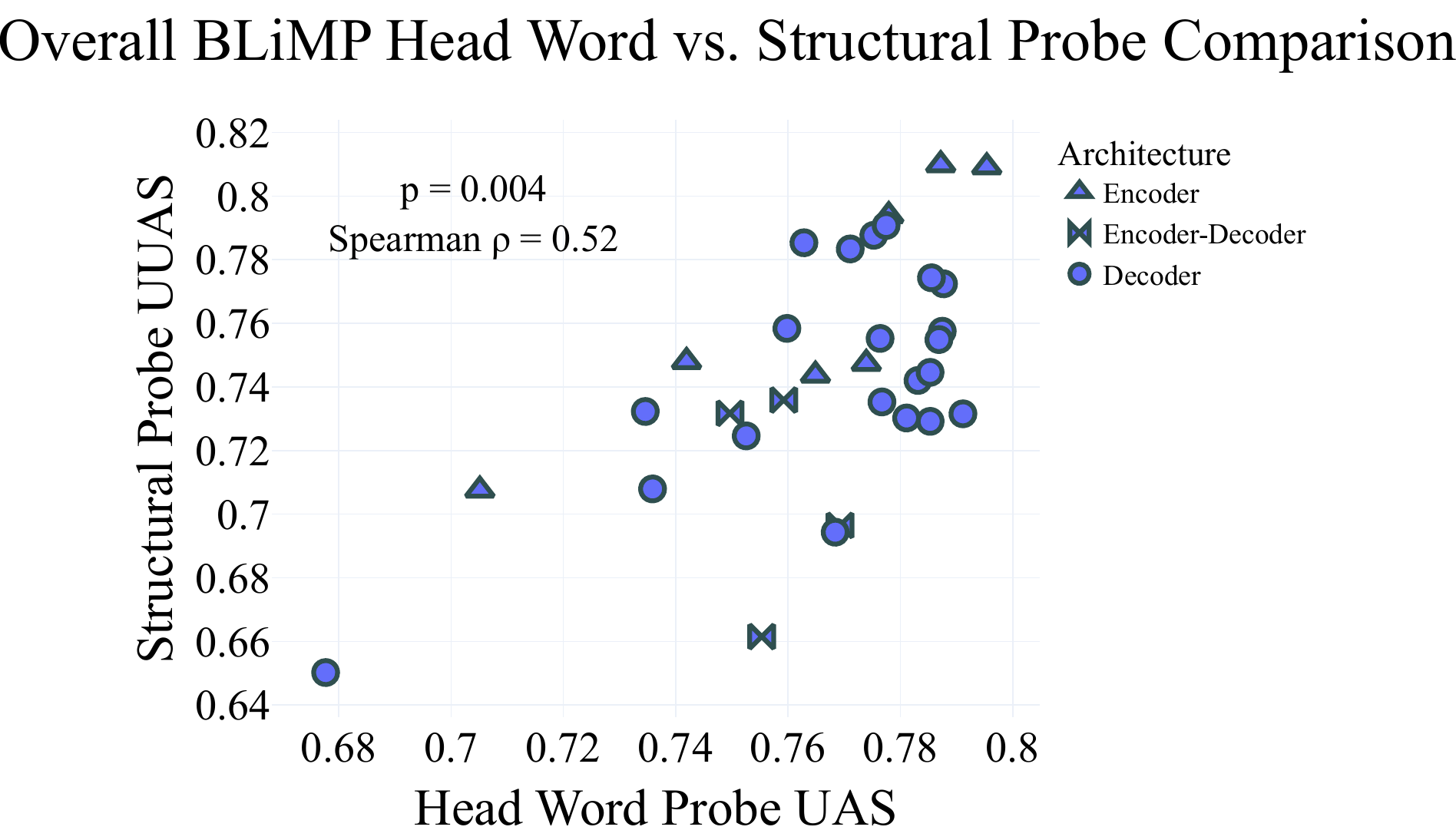}
  \caption{\textbf{Correlating \head{} and \struct{} BLiMP attachment scores.} Across all models, \head{} and \struct{} syntax probes show a significant, moderately positive correlation in accuracy averaged across BLiMP paradigms. Despite having an additional directional aspect, \head{} trains similarly to \struct{}.}
  \label{fig:head_struct_comparison}
\end{figure}

\begin{figure}[!htbp]
\centering
  \includegraphics[width=0.85\textwidth]{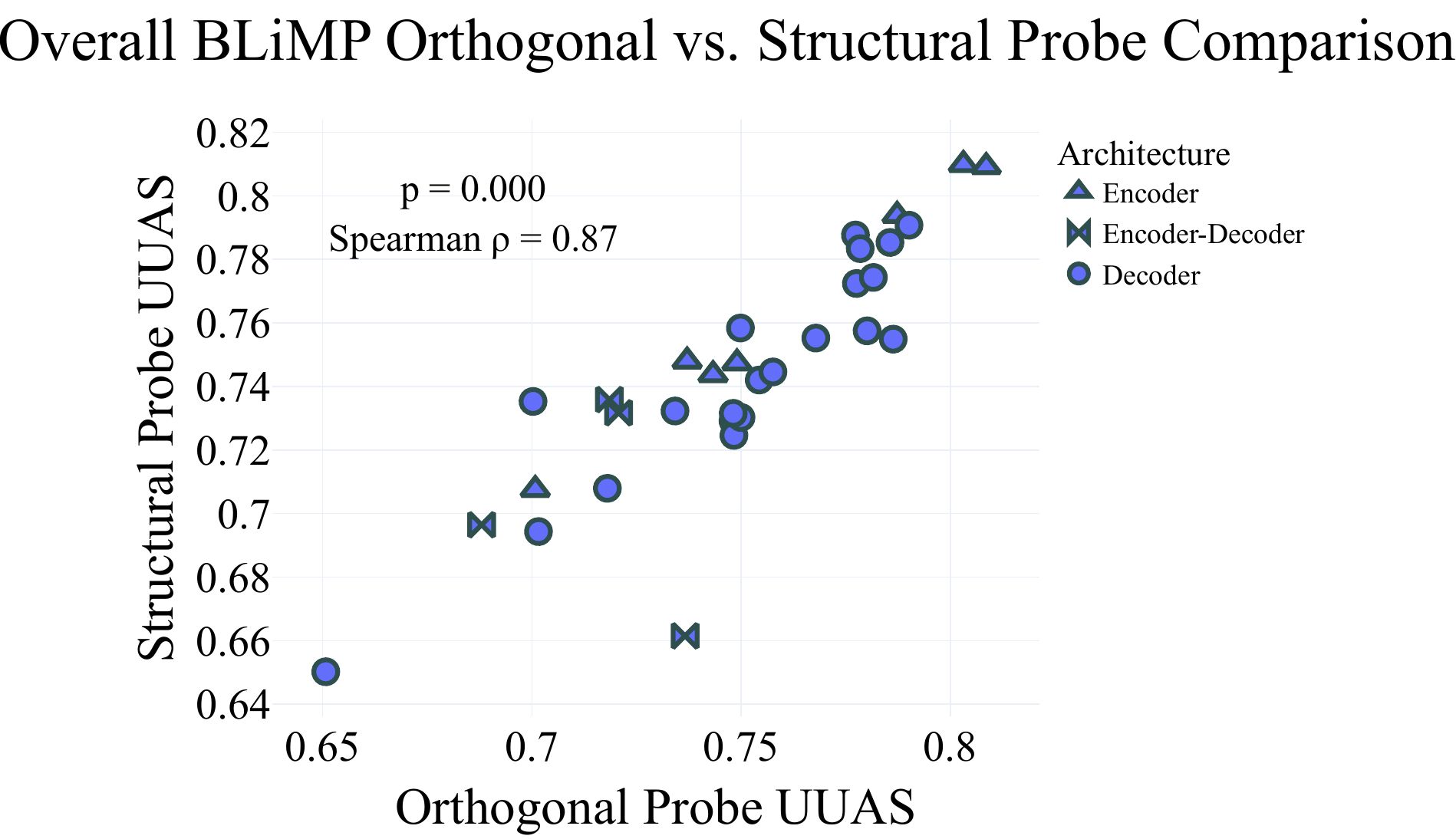}
  \caption{\textbf{Correlating \ortho{} and \struct{} BLiMP attachment scores.} Across all models, \ortho{} and \struct{} syntax probes show a strongly positive correlation in accuracy averaged across BLiMP paradigms. Despite the larger capacity of the orthogonal probe, performance largely mirrors its simpler variant, suggesting that the extra dimensions are not critical to learning the dependency tree probing task.}
  \label{fig:ortho_struct_comparison}
\end{figure}

\FloatBarrier
\clearpage

\section{Full BLiMP and per-Phenomenon OLS Regressions}
\label{sec:full-tables}

\begin{table}[ht]
\centering
\begin{subtable}{0.95\textwidth}
\small
\begin{tabular}{cccc|ccc|cc}
\toprule
\multirow{2}{*}{\textbf{Dataset}} & 
\multicolumn{3}{c|}{\textbf{Simple Regression} (\(x_1 = g^{\text{struct}}_{\phi}\))} & 
\multicolumn{3}{c|}{\textbf{Multiple Regression} (\(x_1, x_2 = g^{\text{ctrl}}_{\phi}\))} & 
\multicolumn{2}{c}{\textbf{LRT}} \\
\cmidrule(lr){2-4} \cmidrule(lr){5-7} \cmidrule(lr){8-9}
 & \(\beta_1\) & $p$-value & Adj.\(R^2\) & \(\beta_1\) & $p$-value & Adj.\(R^2\) & Stat & $p$-value \\
\midrule
Overall & 0.133 & 0.117 & 0.049 & 0.109 & 0.214 & 0.051 & 1.161 & 0.281 \\
Anaphor Agreement & 0.165 & \textbf{0.019} & 0.268 & 0.168 & \textbf{0.040} & 0.243 & 0.027 & 1.0 \\
Binding & 0.398 & 0.106 & 0.178 & 0.386 & 0.155 & 0.155 & 0.18 & 1.0 \\
Ellipsis & -0.421 & 0.106 & 0.181 & -0.426 & 0.152 & 0.153 & 0.011 & 1.0 \\
NPI Licensing & 0.408 & 0.121 & 0.165 & 0.404 & 0.155 & 0.143 & 0.266 & 1.0 \\
S-Selection & -0.249 & 0.242 & 0.125 & -0.213 & 0.374 & 0.255 & 6.254 & 0.149 \\
Island Effects & 0.483 & 0.45 & 0.087 & 0.435 & 0.862 & 0.065 & 0.326 & 1.0 \\
Filler--Gap Dependency & -0.028 & 1.0 & -0.033 & 0.033 & 1.0 & -0.05 & 0.546 & 1.0 \\
Quantifiers & 0.267 & 1.0 & 0.008 & 0.253 & 1.0 & -0.019 & 0.232 & 1.0 \\
Determiner Noun Agr & -0.063 & 1.0 & -0.013 & -0.062 & 1.0 & -0.046 & 0.067 & 1.0 \\
Argument Structure & 0.198 & 1.0 & 0.006 & 0.164 & 1.0 & 0.043 & 2.295 & 1.0 \\
Control/Raising & 0.118 & 1.0 & 0.003 & 0.101 & 1.0 & 0.009 & 1.295 & 1.0 \\
Subject--Verb Agr & -0.108 & 1.0 & 0.007 & -0.047 & 1.0 & 0.128 & 5.236 & 0.243 \\
Irregular Forms & -0.046 & 1.0 & -0.029 & -0.027 & 1.0 & 0.277 & 12.376 & \textbf{0.006}\\
\bottomrule
\end{tabular}
\caption{$x_1 = $ \struct{} UUAS.}
\label{tab:struct_full_table}
\end{subtable}

\vspace{0.5em}

\begin{subtable}{0.95\textwidth}
\small
\begin{tabular}{cccc|ccc|cc}
\toprule
\multirow{2}{*}{\textbf{Dataset}} & 
\multicolumn{3}{c|}{\textbf{Simple Regression} (\(x_1 = g^{\text{ortho}}_{\phi}\))} & 
\multicolumn{3}{c|}{\textbf{Multiple Regression} (\(x_1, x_2 = g^{\text{ctrl}}_{\phi}\))} & 
\multicolumn{2}{c}{\textbf{LRT}} \\
\cmidrule(lr){2-4} \cmidrule(lr){5-7} \cmidrule(lr){8-9}
 & \(\beta_1\) & $p$-value & Adj.\(R^2\) & \(\beta_1\) & $p$-value & Adj.\(R^2\) & Stat & $p$-value \\
\midrule
Overall & 0.127 & 0.157 & 0.035 & 0.102 & 0.266 & 0.041 & 1.299 & 0.254 \\
Anaphor Agreement & 0.173 & \textbf{0.002} & 0.364 & 0.173 & \textbf{0.005} & 0.342 & 0.004 & 1.0 \\
Binding & 0.431 & 0.08 & 0.195 & 0.419 & 0.12 & 0.173 & 0.222 & 1.0 \\
Ellipsis & -0.381 & 0.297 & 0.124 & -0.376 & 0.414 & 0.095 & 0.01 & 1.0 \\
NPI Licensing & 0.326 & 0.618 & 0.082 & 0.32 & 0.715 & 0.057 & 0.215 & 1.0 \\
S-Selection & -0.216 & 0.727 & 0.068 & -0.175 & 1.0 & 0.205 & 6.164 & 0.156 \\
Filler--Gap Dependency & -0.1 & 1.0 & -0.027 & -0.058 & 1.0 & -0.049 & 0.402 & 1.0 \\
Quantifiers & 0.297 & 1.0 & 0.015 & 0.266 & 1.0 & -0.012 & 0.21 & 1.0 \\
Determiner Noun Agr & -0.089 & 1.0 & -0.004 & -0.089 & 1.0 & -0.036 & 0.077 & 1.0 \\
Argument Structure & 0.174 & 1.0 & -0.003 & 0.132 & 1.0 & 0.032 & 2.24 & 1.0 \\
Control/Raising & 0.165 & 1.0 & 0.041 & 0.153 & 1.0 & 0.048 & 1.304 & 1.0 \\
Subject--Verb Agr & -0.119 & 1.0 & 0.014 & -0.059 & 1.0 & 0.132 & 5.183 & 0.251 \\
Island Effects & 0.42 & 1.0 & 0.045 & 0.356 & 1.0 & 0.028 & 0.493 & 1.0 \\
Irregular Forms & -0.093 & 1.0 & -0.016 & -0.078 & 1.0 & 0.287 & 12.45 & \textbf{0.005} \\
\bottomrule
\end{tabular}
\caption{$x_1 = $ \ortho{} UUAS.}
\label{tab:regression_ortho_results}
\end{subtable}

\vspace{0.5em}


\begin{subtable}{0.95\textwidth}
\small
\begin{tabular}{cccc|ccc|cc}
\toprule
\multirow{2}{*}{\textbf{Dataset}} & 
\multicolumn{3}{c|}{\textbf{Simple Regression} (\(x_1 = g^{\text{head}}_{\phi}\))} & 
\multicolumn{3}{c|}{\textbf{Multiple Regression} (\(x_1, x_2 = g^{\text{ctrl}}_{\phi}\))} & 
\multicolumn{2}{c}{\textbf{LRT}} \\
\cmidrule(lr){2-4} \cmidrule(lr){5-7} \cmidrule(lr){8-9}
 & \(\beta_1\) & $p$-value & Adj.\(R^2\) & \(\beta_1\) & $p$-value & Adj.\(R^2\) & Stat & $p$-value \\
\midrule
Overall & 0.0 & 0.998 & -0.033 & -0.048 & 0.700 & 0.091 & 5.194 & \textbf{0.023} \\
Binding & 0.534 & 0.064 & 0.209 & 0.522 & 0.108 & 0.184 & 0.085 & 1.0 \\
S-Selection & -0.2 & 1.0 & -0.007 & -0.255 & 1.0 & 0.068 & 3.553 & 0.654 \\
NPI Licensing & 0.25 & 1.0 & 0.006 & 0.233 & 1.0 & -0.001 & 0.882 & 1.0 \\
Filler--Gap Dependency & -0.125 & 1.0 & -0.028 & -0.15 & 1.0 & -0.055 & 0.256 & 1.0 \\
Quantifiers & 0.434 & 1.0 & 0.045 & 0.385 & 1.0 & 0.1 & 2.98 & 0.843 \\
Determiner Noun Agr & -0.096 & 1.0 & -0.013 & -0.098 & 1.0 & -0.046 & 0.037 & 1.0 \\
Ellipsis & -0.109 & 1.0 & -0.02 & -0.1 & 1.0 & -0.053 & 0.087 & 1.0 \\
Argument Structure & 0.025 & 1.0 & -0.033 & -0.037 & 1.0 & 0.076 & 4.658 & 0.402 \\
Control/Raising & 0.241 & 1.0 & 0.047 & 0.201 & 1.0 & 0.076 & 2.083 & 1.0 \\
Subject--Verb Agr & -0.205 & 1.0 & 0.041 & -0.221 & 1.0 & 0.025 & 0.56 & 1.0 \\
Island Effects & 0.392 & 1.0 & -0.001 & 0.397 & 1.0 & -0.035 & 0.025 & 1.0 \\
Anaphor Agreement & 0.102 & 1.0 & 0.005 & 0.096 & 1.0 & -0.027 & 0.069 & 1.0 \\
Irregular Forms & -0.120 & 1.0 & -0.026 & -0.099 & 1.0 & 0.069 & 4.217 & 0.48 \\
\bottomrule
\end{tabular}
\caption{$x_1 = $ \head{} UAS. That 12 out of 13 phenomena have $p$-value 1.0 is a striking result.}
\label{tab:regression_head_results}
\end{subtable}

\caption{\textbf{Comparison of simple and multiple regression statistics for all BLiMP linguistic phenomena.} Per-phenomenon rows are sorted in order of increasing simple regression corrected $p$-value. Phenomena show substantial variation in strength of association. Zeroing into Subtable~\ref{tab:struct_full_table}, we note the adjusted $R^2$ jumps for s-selection and irregular forms after adding the control. The significant $p$-value of the irregular forms LRT concurs with Fig~\ref{fig:irregular_forms_control}.}
\label{tab:regression-comparison-full}
\end{table}


\begin{table}[!htbp]
\centering
\small
\begin{tabular}{cccc}
\toprule
\multirow{2}{*}{\textbf{Dataset}} & 
\multicolumn{3}{c}{\textbf{Simple Regression} ($x_1 = g^{\text{ctrl}}_{\phi}$)} \\
\cmidrule(lr){2-4}
& $\beta_1$ & $p$-value & Adj.\ $R^2$ \\
\midrule
Overall & 0.127 & 0.165 & 0.032 \\
Irregular Forms & 0.287 & \textbf{0.009} & 0.299 \\
S-Selection & -0.445 & 0.141 & 0.167 \\
Subject--Verb Agr & -0.343 & 0.178 & 0.151 \\
Binding & 0.118 & 1.0 & -0.009 \\
NPI Licensing & 0.086 & 1.0 & -0.022 \\
Filler--Gap Dependency & -0.193 & 1.0 & -0.016 \\
Quantifiers & 0.232 & 1.0 & -0.015 \\
Determiner Noun Agr & -0.036 & 1.0 & -0.031 \\
Ellipsis & -0.134 & 1.0 & -0.018 \\
Argument Structure & 0.208 & 1.0 & 0.048 \\
Control/Raising & 0.161 & 1.0 & 0.016 \\
Island Effects & 0.202 & 1.0 & 0.01 \\
Anaphor Agreement & 0.025 & 1.0 & 0.005 \\
\bottomrule
\end{tabular}
\caption{\textbf{Simple linear regression results for the structural probe control (trained on the best layers for \struct{}) $x_1 = $ \control{} $\rho_s$}. Per-phenomenon rows are sorted in order of increasing corrected $p$-value. Irregular forms has statistical significance and substantially higher $R^2$ than most of the syntax probe simple regressions for any phenomena (Table~\ref{tab:struct_full_table} and Table~\ref{tab:regression_head_results}). This critical result shows that even a non-syntactic control can achieve a significant positive association with one syntactic benchmark, which thus urges caution when interpreting the statistical signifiance of anaphor agreement for \struct{}.}
\label{tab:simple_ctrl_regression}
\end{table}

\begin{table}[!htbp]
\centering
\small
\begin{tabular}{cccc}
\toprule
\multicolumn{1}{c}{\textbf{Dataset}} & 
\multicolumn{3}{c}{\textbf{Simple Regression} ($x_1 = g^{\text{ctrl}}_{\phi}$)} \\
\cmidrule(lr){2-4}
& $\beta_1$ & $p$-value & Adj.\ $R^2$ \\
\midrule
Overall & 0.043 & \textbf{0.031} & 0.117 \\
Argument Structure & 0.085 & 0.5 & 0.106 \\
Irregular Forms & 0.078 & 0.568 & 0.096 \\
Quantifiers & 0.143 & 0.865 & 0.069 \\
S-Selection & 0.043 & 1.0 & 0.057 \\
Binding & 0.038 & 1.0 & -0.007 \\
NPI Licensing & 0.038 & 1.0 & -0.001 \\
Filler--Gap Dependency & 0.027 & 1.0 & -0.027 \\
Determiner Noun Agr & 0.002 & 1.0 & -0.033 \\
Ellipsis & -0.023 & 1.0 & -0.028 \\
Control/Raising & 0.049 & 1.0 & 0.053 \\
Subject--Verb Agr & 0.008 & 1.0 & -0.026 \\
Island Effects & -0.004 & 1.0 & -0.033 \\
Anaphor Agreement & 0.004 & 1.0 & -0.025 \\
\bottomrule
\end{tabular}
\caption{\textbf{Simple linear regression results for the head word probe control (trained on the best layers for \head{}) $x_1 = $ \control{} $\rho_s$}. Per-phenomenon rows are sorted in order of increasing simple regression corrected $p$-value. While all phenomena have extremely low or near-zero explanatory power that does not achieve statistical significance, the fit on the full dataset is significant but has near-zero effect size.}
\label{tab:simple_ctrl_regression_head}
\end{table}


\FloatBarrier

\section{Subject--Verb Agreement \& Filler--Gap Experiments}
\label{sec:full-hamming}

\begin{figure}[htbp]
    \includegraphics[width=0.95\textwidth]{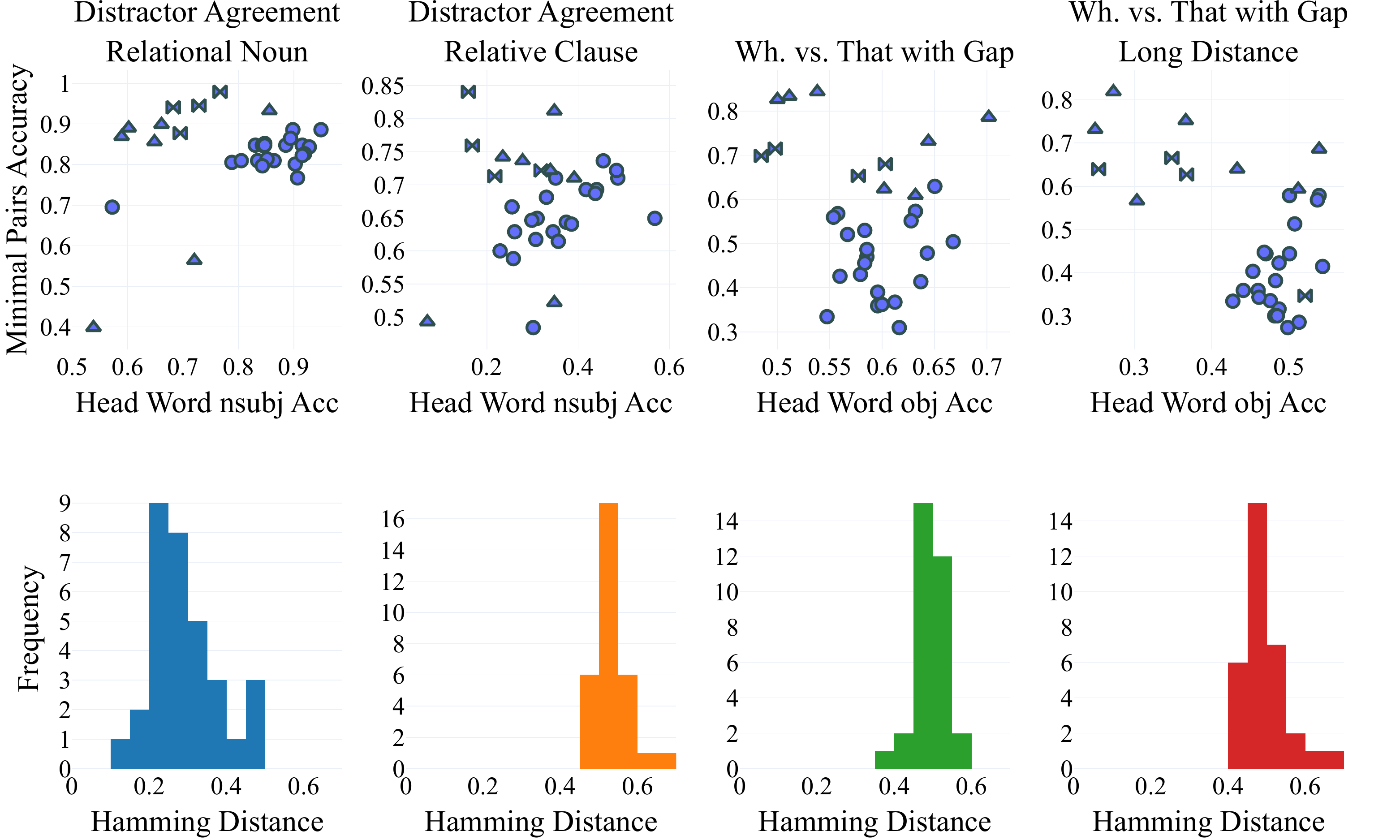}
    \caption{Analogous to Fig~\ref{fig:struct_hamming_details} but for the head word probe. Conclusion is similar.}
    \label{fig:head_hamming_details}
\end{figure}

\end{document}